\documentclass[journal]{IEEEtran}
\usepackage{cite}
\usepackage{color}
\usepackage{amsmath,amssymb,amsfonts, mathrsfs, bm, mathabx}
\usepackage{multirow}
\usepackage{algorithmic}
\usepackage{hyperref}
\usepackage{graphicx}
\usepackage{textcomp}
\usepackage{tabularx,booktabs}
\usepackage{stfloats}
\usepackage{longtable}
\usepackage[ruled]{algorithm2e}

\usepackage{url}
\usepackage{supertabular,booktabs}
\usepackage{array}
\usepackage{verbatim}
\usepackage{enumitem}
\newcommand {\mymarginpar}[1]{\marginpar{#1}}
\renewcommand {\marginpar}[1]{}
\def\_{\rule{.3em}{.15ex}}      % Get underscore by typing \_.

\newcommand {\bearn}{\begin{eqnarray*}}
\newcommand {\eearn}{\end{eqnarray*}}
\newcommand {\barr}{\begin{array}}
\newcommand {\earr}{\end{array}}

\renewcommand {\L}{{\cal L}}

\newcommand {\N}{{\cal N}}

\newtheorem{definition}{Definition}
\newtheorem{property}[definition]{Property}
\newtheorem{proposition}[definition]{Proposition}
\newtheorem{lemma}[definition]{Lemma}
\newtheorem{theorem}[definition]{Theorem}
\newtheorem{corollary}[definition]{Corollary}
\newtheorem{example}{Example}
\newtheorem{remark}[definition]{Remark}

\newcommand {\benum} {\begin{enumerate}}
\newcommand {\eenum} {\end{enumerate}}
\newcommand {\bdesc} {\begin{description}}
\newcommand {\edesc} {\end{description}}

\newcommand {\bfig}[2] {
    \begin{figure}
    \centering
    \includegraphics[width=#2]{#1}
}
\newcommand {\brotatefig}[2]{
    \begin{figure}[htbp]
    \centerline{\epsfig{figure={#1},clip=,angle=-90,width={#2}}}
}
\newcommand {\bfigfirst}[2] {\begin{figure}[h]
                        \centerline {
                        \setlength{\epsfxsize}{#2}
                        \epsffile{#1}}}
\newcommand {\efig}[2]{ \caption{#2}
                        \label{fig:#1}
                        \end{figure}
                        \mymarginpar{fig:#1}}
\newcommand {\erotatefig}[2]{ \caption{#2}
                        \label{fig:#1}
                        \end{figure}
                        \mymarginpar{fig:#1}}
\newcommand {\rfig}[1]{Figure \ref{fig:#1}}

\newcommand {\btab}[1]{
                       \begin{table}
                       \centering
                       \begin{tabular}{#1}}
\newcommand {\etab}[3] {
                       \end{tabular}
                       \caption[#3]{#2}
                       \label{tab:#1}
                       \end{table}
                       \mymarginpar{tab:#1}
                       \vspace{.1in}}

\newcommand {\btabular}[1]{\begin{center}
                       \begin{tabular}{#1}}
\newcommand {\etabular}{\end{tabular}
                       \end{center}}

\newcommand {\bdefin}[1]{\begin{definition}
                      \mymarginpar{def:#1}
                      \label{def:#1} }
\newcommand {\edefin}       {\end{definition}}
\newcommand {\rdef}[1]{Definition \ref{def:#1}}

\newcommand {\bpro}[1]{\begin{property}
                      \mymarginpar{pro:#1}
                      \label{pro:#1} }
\newcommand {\epro}   {\end{property}}

\newcommand {\bprop}[1]{\begin{proposition}
                      \mymarginpar{prop:#1}
                      \label{prop:#1} }
\newcommand {\eprop}       {\end{proposition}}
\newcommand {\rprop}[1]{Proposition \ref{prop:#1}}

\newcommand {\blem}[1]{\begin{lemma}
                      \mymarginpar{lem:#1}
                      \label{lem:#1} }
\newcommand {\elem}   {\end{lemma}}

\newcommand {\bthe}[1]{\begin{theorem}
                      \mymarginpar{the:#1}
                      \label{the:#1} }
\newcommand {\ethe}   {\end{theorem}}

\newcommand {\bcor}[1]{\begin{corollary}
                      \mymarginpar{cor:#1}
                      \label{cor:#1} }
\newcommand {\ecor}   {\end{corollary}}

\newcommand {\bax}[1]{\begin{axiom}
                      \mymarginpar{ax:#1}
                      \label{ax:#1} }
\newcommand {\eax}       {\vspace{-.1in} \end{axiom}}

\newcommand {\bex}[2]{\vspace{.1in}
                      \begin{example}
                      \mymarginpar{ex:#1}
                       {\bf #2}
                      \label{ex:#1} }
\newcommand {\eex}       {\end{example} \vspace{.3cm} }

\newcommand {\brem}[1]{\begin{remark}
                      \mymarginpar{rem:#1}
                      \label{rem:#1} \em }
\newcommand {\erem}   {\end{remark}}

\newcommand {\beq}[1]{\mymarginpar{eq:#1}
                      \begin{equation}
                      \label{eq:#1} }

\newcommand {\beqno}[1]{\mymarginpar{eq:#1}
                      \begin{eqnarray}
                      \nonumber}

\newcommand {\eeq}       {\end{equation}}
\newcommand {\eeqno}       { && \end{eqnarray}}
\newcommand {\req}[1]{(\ref{eq:#1})}

\newcommand {\bear}[1]{\mymarginpar{eq:#1}
                       \begin{eqnarray}
                       \label{eq:#1} }

\newcommand {\bearno}[1]{\mymarginpar{eq:#1}
                       \begin{eqnarray}
                       \nonumber}

\newcommand {\eear}{\end{eqnarray}}
\newcommand {\eearno}{\end{eqnarray}}
\newcommand {\bsel}{\left \{ \begin{array}{cl}}
\newcommand {\esel}{\end{array} \right.}

\newcommand {\bmat}[1]{\left [ \begin{array}{#1}}
\newcommand {\emat}{\end{array} \right ]}
\newcommand {\bsec}[2]{\mymarginpar{sec:#2}
                       \section{#1}
                       \label{sec:#2} }

\newcommand {\rsec}[1]{Section \ref{sec:#1}}

\newcommand {\bsubsec}[2]{\mymarginpar{sec:#2}
                       \subsection{#1}
                       \label{sec:#2} }

\def\R{I\kern-0.30em R}
\def\N{I\kern-0.30em N}
\def\P{I\kern-0.30em P}

\def\pr{{\bf\sf P}}

\newcommand{\rhog}{\rho}

\def\calS{{\cal S}}
\def\calT{{\cal T}}
\def\calD{{\cal D}}

\def\red{\color{red}}

\definecolor{darkgreen}{rgb}{0.0, 0.5, 0.0}
\def\yw{\color{black}}
\def\cs{\color{black}}

\begin{document}

\title{A Mathematical Theory for Learning Semantic Languages by Abstract Learners}

\author{Kuo-Yu Liao,
        Cheng-Shang~Chang,~\IEEEmembership{Fellow,~IEEE,}\\
		and Y.-W. Peter Hong,~\IEEEmembership{Senior Member,~IEEE}
		\thanks{The authors are with the Institute of Communications Engineering, National Tsing Hua University, Hsinchu 300044, Taiwan R.O.C. Email: d25602685@gmail.com; cschang@ee.nthu.edu.tw; ywhong@ee.nthu.edu.tw.}
\thanks{%Manuscript received November 17, 2020; revised March 29, 2021; accepted May 16, 2021.
 This work was supported in part by the Ministry of Science and Technology, Taiwan, under Grant 111-2221-E-007-038-MY3 and 111-2221-E-007-042-MY3, and in part by Qualcomm Technologies under Grant SOW NAT-487844-2.}
 % (Corresponding author: Y.-W. Peter Hong.)}

 \vspace{-.5cm}
 }

\maketitle
\thispagestyle{empty}

\begin{abstract}
Recent advances in Large Language Models (LLMs) have demonstrated the emergence of capabilities (learned skills) when the number of system parameters and the size of training data surpass certain thresholds. The exact mechanisms behind such phenomena are not fully understood and remain a topic of active research. Inspired by the skill-text bipartite graph model proposed by Arora and Goyal
for modeling semantic languages, we develop a mathematical theory to explain the emergence of learned skills, taking the learning (or training) process into account. Our approach models the learning process for skills in the skill-text bipartite graph as an iterative decoding process in Low-Density Parity Check (LDPC) codes and Irregular Repetition Slotted ALOHA (IRSA). Using density evolution analysis, we demonstrate the emergence of learned skills when the ratio of the number of training texts to the number of skills exceeds a certain threshold. Our analysis also yields a scaling law for testing errors relative to this ratio. Upon completion of the training,
the association of learned skills can also be acquired to form a skill association graph. We use site percolation analysis to derive the conditions for the existence of a giant component in the skill association graph. Our analysis can also be extended to the setting with
a hierarchy of skills, where a fine-tuned model is built upon a foundation model. It is also applicable to the setting with multiple classes of skills and texts. As an important application,
we propose a method for semantic compression and discuss its connections to semantic communication.
 \end{abstract}

%\iffalse
\begin{IEEEkeywords}
Large language models, emergence of capabilities, Low-Density Parity Check codes, Irregular Repetition Slotted ALOHA, density evolution, semantic communication
\end{IEEEkeywords}
%\fi

%%%%%%%%%%%%%%%%%%%%%%%%%%%%%%%%%%%%%%%%%%%%%%%%%%
\bsec{Introduction}{introduction}
%%%%%%%%%%%%%%%%%%%%%%%%%%%%%%%%%%%%%%%%%%%%%%%%%%

In the recent era of natural language processing (NLP), the evolution of large language models (LLMs), such as GPT-4 \cite{GPT4} and Gemini \cite{pichai2023introducing}, has greatly impacted people's daily lives. There is a growing consensus that enhancing language model performance and sample efficiency across a broad spectrum of downstream NLP tasks is closely linked to scaling up these models. This scaling involves increasing both the size of the training data and the number of model parameters, as discussed in \cite{devlin2018bert,brown2020language}. The relationship between the scale of an LLM and its performance can often be quantitatively forecasted using scaling laws,
as explored
in \cite{kaplan2020scaling,hoffmann2022training}.

Recent studies \cite{wei2022emergent,wei2022inverse,GPT4} further demonstrates, through various numerical examples, that LLMs manifest emergent capabilities absent in their smaller counterparts. There is a notable enhancement in system performance once a certain critical scale threshold is surpassed, exhibiting a {\em phase transition} behavior often observed in network science \cite{Newman2010}.
However, as pointed out in \cite{wei2022emergent}, the exact mechanisms behind
these emergent abilities in large language models are still not fully understood and remain a topic of active research.

In the study presented by \cite{chang2023simple}, a simple explanation is offered for the phase transition phenomenon observed in LLMs based on the concept of list decoders. The approach in \cite{chang2023simple} models an LLM as a sequence-to-sequence random function over a certain token space with $M$ possible tokens. It was shown that the expected number of erroneous sequences in the list decoder can be bounded by a constant if $M\epsilon < 1$, where $\epsilon$ is the false alarm probability. Since transformer-based LLMs with more parameters and extensive training can memorize more patterns \cite{vaswani2017attention,ramsauer2020hopfield}, they are more likely to reduce the false alarm probability $\epsilon$ below the percolation threshold of $1/M$. However, the list decoder approach does not explain how the capabilities of language skills emerge from scaling up
the size of the training data. In \cite{arora2023theory}, a novel random graph approach was proposed to explain the scaling law discussed in \cite{hoffmann2022training}. A language is modeled by a random bipartite graph with {\em skills} on one side and {\em texts} (originally termed {\em text-pieces} in \cite{arora2023theory}) on the other side. An edge between a text and a skill indicates that understanding the text requires the skill. A text can be comprehended if all the skills connected to it are acquired.

Inspired by the skill-text bipartite graph model proposed in \cite{arora2023theory}, we develop a mathematical theory to explain the emergence of learned skills by an abstract learner during the {\em learning} (or {\em training}) process. We summarize our contributions as follows:

\begin{itemize}[itemindent=0pt]
\item[(i)]  {\bf Learning as an Iterative Decoding Process:}
We model the {\em learning/training} process for the skills in the skill-text bipartite graph as an iterative decoding process, similar to {\yw that adopted} in Low-Density Parity Check (LDPC) codes \cite{gallager1962low,shokrollahi1999new,richardson2001design} and  Irregular Repetition Slotted ALOHA (IRSA) \cite{liva2011graph,narayanan2012iterative,paolini2012random,jakovetic2015cooperative,stefanovic2018coded,chiang2022parallel}. When a text is presented to a learner, a skill in that text might be learned with a certain probability (specific definitions for learners are given in \rsec{learners}). By repeatedly presenting a large number of training texts to a learner, a fraction of skills can be learned.

\item[(ii)] {\bf The Emergence of Learned Skills:} Following the density evolution analysis for the iterative decoding process (in the asymptotic regime of a large number of skills) \cite{luby1998analysis,luby1998analysisb,richardson2001capacity}, we show the emergence of learned skills when the ratio $R$ of the number of training texts to the number of skills exceeds a certain threshold. Once this threshold is exceeded, the testing error, defined as the probability {\yw that a randomly selected text is not understood by the learner,}
        drops sharply. Our analysis also provides the scaling law of the testing error with respect to the ratio $R$.

\item[(iii)] {\bf The Association of Skills:} Upon completing training, a learner acquires not just
isolated skills but also {\yw the interrelation of these skills,}
which can be represented by a {\em skill association graph} where edges connect learned skills appearing in the same text. Understanding this graph is crucial for making inferences, such as predicting the next skill from a given set. This process is akin to prompting a language model to forecast and generate a text based on a skill set. This leads to a critical inquiry: Are the learned skills sufficiently interconnected to aid in prediction? To address this question, we use site (or node) percolation analysis \cite{Newman2010} to derive conditions for the existence of a giant component in the skill association graph.

\item[(iv)] {\bf Hierarchy of Skills:}
We consider a setting with two classes of skills: {\em basic} and {\em domain-specific}. Learning a domain-specific skill first requires acquiring a random number of basic skills, establishing a hierarchy of skills. Initially, we train a foundation model on basic skills, which is then fine-tuned with domain-specific texts. Our findings identify two critical thresholds for minimizing testing errors in domain-specific tasks: (i) a large number of basic skills are learned when the number of basic texts
exceeds the threshold in the foundation model, and (ii) a large number of domain-specific skills are learned when the
number of domain-specific texts exceeds the threshold in the fine-tuning model.

\item[(v)] {\bf Multiple Classes of Skills and Texts:}  We extend the density evolution analysis to the settings with multiple classes of skills and texts, driven by the diversity of subjects in texts like math, physics, chemistry, law, etc. Such an extension is rooted on the framework of Poisson receivers with multiple classes of users and receivers in \cite{chang2022stability}. Interestingly, thresholds are also observed for the setting with multiple classes of skills and texts.  The thresholds for various skill classes align due to the coupling of the density evolution equations, as illustrated in \cite{chang2022stability}, resulting in the {\em simultaneous} emergence of various classes of skills.

\item[(vi)] {\bf Application to Semantic Compression:} We propose a semantic compression scheme that encodes the learned skills instead of all words in a text. In particular, we define the notion of a {\em generative} learner that is trained to identify the set of skills in a semantic language. Once trained, it can be used to perform semantic compression by representing a text using indices of the learned skills required to understand that text.

\end{itemize}

The paper is organized as follows. In \rsec{semantic}, we define a semantic language using a skill-text bipartite graph. In \rsec{learners}, we introduce abstract learners, including 1-skill learners in \rsec{one} and Poisson learners in \rsec{Poisson}. Density evolution analysis is employed to derive the scaling law for these abstract learners. In \rsec{skillclusters}, we address the problem of learning the \emph{association} between two learned skills and derive conditions for the existence of a giant component in the skill association graph. In \rsec{hskills}, we further address the problem setting with a hierarchy of skills by using a foundation model and a fine-tuning model. We present the extension to the setting with multiple classes of skills and texts in \rsec{ES}. In \rsec{compression}, we demonstrate how the trained learners can be utilized for semantic compression and semantic communication. The paper concludes in \rsec{con}, where we discuss possible extensions of our work.

\bsec{Semantic Languages}{semantic}

\bsubsec{Definition of a Semantic Language}{defsem}

In this section, we define the notion of semantic languages. This definition is inspired by the recent skill-text bipartite graph model presented in \cite{arora2023theory}, which explains the emergence of new skills in language models.

\bdefin{language}({\bf Semantic Language})
A {\em semantic language} ${\cal L}=({\cal A},{\yw \calT,\calS,} \phi)$ consists of (i) a set of tokens (symbols) ${\cal A}$, (ii) a set of texts $\yw \calT$ composed of sequences of tokens, (iii) a set of skills $\yw\calS$, and (iv) a function $\phi$ that maps a text $\yw t\in\calT$ to a set of {\em skills} $\yw\phi(t)\subseteq\calS$.
\edefin

\begin{figure}[t]
    \centering
    \includegraphics[width=0.8\columnwidth]{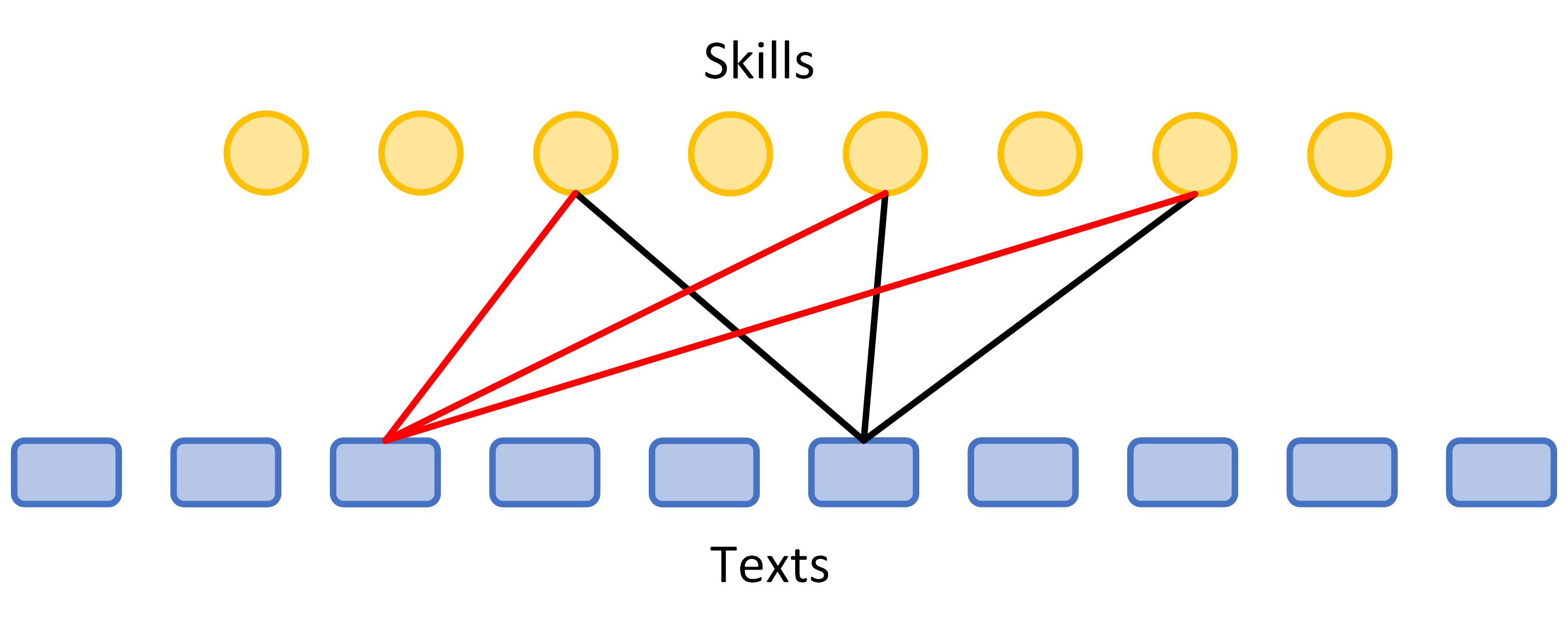}
    \caption{An illustration for a skill-text bipartite graph, where the skill nodes connected to a text node indicates the set of skills required to understand the text.
     Two texts requiring the same set of skills are semantically equivalent.}	
    \label{fig:skilltext}
    \end{figure}

Two texts $t_1$ and $t_2$ are said to be {\em semantically equivalent} (or simply equivalent) if they both require the same set of skills, i.e., $\phi(t_1)=\phi(t_2)$. We say that a skill $s$ is present in a text $t$ if $s$ is included in the set of skills $\phi(t)$. For a semantic language ${\cal L}=({\cal A},{\yw \calT,\calS,}\phi)$, we can generate a skill-text bipartite graph $G=({\yw \calS,\calT,} E)$, where an edge is added between a skill $s$ and a text $t$ if $s$ is present in $\phi(t)$ (see \rfig{skilltext} for an illustration). Note that the skill-text bipartite graph $G=({\yw \calS,\calT,} E)$ characterizes the semantic language.

To motivate the definition of a semantic language, let us consider English as an example. The set of tokens in English comprises the characters that can be produced on a computer keyboard, typically encoded by the standard ASCII code. The set of English texts is vast, containing, as a subset, all the English texts available on the Internet. The set of skills required to understand these texts can also be very large and are generally {\em latent} and difficult to characterize.

In this paper, we address the question of learning a semantic language. A common approach is to sample a subset of texts from the set of texts $\calT$ and present this subset to a {\em learner} to learn the mapping $\phi$.

\bdefin{learned}
A skill $s$ in a semantic language is {\em learned} if the learner can determine whether the skill $s$ is present in any given text $t$ in $\yw\calT$.
A text $t$ is {\em understood} by a learner if all the skills contained within $t$ are learned by the learner.
\edefin

\bsubsec{Learning a Semantic Language by Sampling}{sampling}

Suppose that a subset  of  texts $\calD$ is sampled from the set of texts $\yw\calT$ and presented to the learner as {\em training texts}. Let $R = |\calD|/|\yw\calS|$ be the ratio of the number of training texts to the number of skills.

We make the following two assumptions about the sampled texts:

\begin{description}
\item [(A1)] (Poisson degree distribution) Let $N(t)=|\phi(t)|$ be the number of skills in the  text $t \in  \calD$.
Then $\{N(t), t \in  \calD\}$ are independent Poisson random variables with mean $c$, i.e., $\pr(N(t) = k) = e^{-c} \frac{c^k}{k!}$ for $k = 0, 1, 2, \ldots$.

\item [(A2)] (Uniform connection) Each edge of a text node is connected to a skill node independently and uniformly. Thus, the probability that an edge of the text node $t$ is connected to a particular skill node $s$ is $1/|\calS|$.
\end{description}

From (A1) and (A2), the skill-text bipartite graph of a {\em sampled} semantic language is a random bipartite graph with $|\calS|$ skill nodes on one side and $|\calD|$ text nodes on the other side.
Denote by $M(s,t)$ the number of edges between a skill node $s$ and a text node $t$.
Since randomly splitting a Poisson process yields {\em independent} Poisson processes, we have from (A1) and (A2) that $\{M(s,t), s \in  \calS, t \in \calD\}$ are independent Poisson random variables
with mean $c/|\calS|$. Denote by $M(s)$ the number of edges of a skill node $s$.
As a superposition of independent Poisson processes (from the $|\calD|$ text nodes) is a Poisson process,  we know that $\{M(s), s \in \calS\}$ are independent Poisson random variables
with mean $c|\calD|/|\calS|$
(i.e., $cR$). On the other hand, the probability that a point in a superposition of independent Poisson processes comes from one of these processes is proportional to their rates.  This leads to the following proposition.

\bprop{skilldegree}
Consider a sampled skill-text bipartite graph from a semantic language ${\cal L}=({\cal A},{\calT,\calS,} \phi)$. Then:
\begin{description}
\item[(i)] {\cs (Poisson degree distribution)  $\{M(s), s \in \calS\}$ are independent Poisson random variables
with mean $c|\calD|/|\calS|=cR$, where $M(s)$ is the degree of the skill node $s$.
\item[(ii)] (Uniform connection) Each edge of a skill node is connected to a text node independently and uniformly. Thus, the probability that an edge of a skill node is connected to a particular text node is $1/|\calD|$.}
\end{description}
Therefore, such a bipartite graph is a random bipartite graph, as per the configuration model in \cite{Newman2010}, where the degree distribution of the text nodes (respectively, skill nodes) is Poisson with mean $c$ (respectively, $c R$).
\eprop

\iffalse 
{\red [Is the randomly selected edge selecting each edge uniformly, or first choosing the text nodes uniformly, then choosing the edge uniformly from those connected to the text node.]} {\cs I've rewritten this by using i.i.d. Poisson random variables. This should be clear how the random bipartite graph is constructed.}
\fi

\bsec{Abstract Learners}{learners}

In this section, we develop a mathematical theory to understand the learning process of a semantic language by an abstract learner. To a learner, all skills are {\em novel} before training, but a skill $s$ in text $t$ may be {\it learned} after the text is presented to the learner.
By repeatedly presenting a set of training texts $\calD$, we aim to determine the fraction of skills that can be learned by the learner. We consider two types of learners based on their learning capabilities, namely, a 1-skill learner and a more general Poisson learner.

\bsubsec{1-Skill Learner}{one}

\bdefin{oneskill} ({\bf 1-Skill Learner}) An abstract learner is called a {\em 1-skill learner} if it can learn a novel skill $s$ by being presented with a text $t$ where the skill $s$ is the only novel skill in the text. Once a skill $s$ is learned, the learner is able to identify the skill $s$ appearing in any other {\yw text.}
\edefin

Now we describe how the 1-skill learner is trained. For each iteration, we present the $|\calD|$ sampled texts to the 1-skill learner in parallel. In the first iteration, a text containing only one skill is used to learn the skill in that text. Texts containing more than one skill do not contribute to learning in the first iteration. Once a skill $s$ is learned by the 1-skill learner, the number of novel skills in other texts containing $s$ is reduced by 1. In other words, the edges connected to the skill node $s$ can be removed from the skill-text bipartite graph. Therefore, the number of novel skills associated with each text in the next iteration will not be greater than that in the current iteration. In the second iteration, texts with only one novel skill remaining are used to learn the skills in those texts. As in the first iteration, skills learned in the second iteration can be used to remove the corresponding edges in the skill-text bipartite graph. This iteration process is repeated until no more novel skills can be learned. We refer to this training process as the {\em Successive Cancellation of Novel Skills (SCNS)}.

\begin{figure}[t]
    \centering
    \includegraphics[width=0.8\columnwidth]{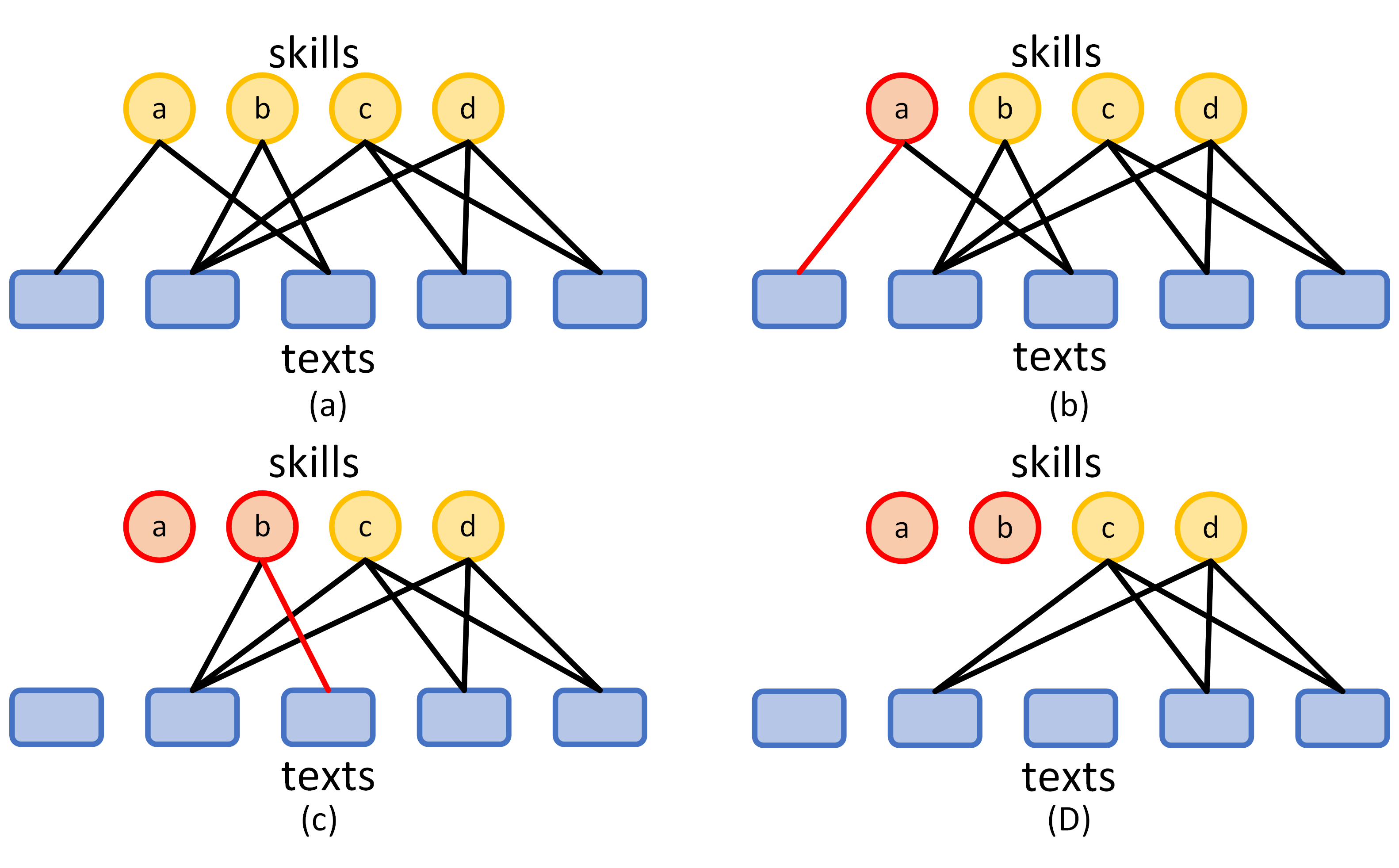}
    \caption{An example of the {\em Successive Cancellation of Novel Skills (SCNS)}.}	
    \label{fig:SCNS}
    \end{figure}

In \rfig{SCNS}, we present a simple example of SCNS. \rfig{SCNS}(a) depicts the original skill-text bipartite graph. \rfig{SCNS}(b) focuses on the skill node $a$, which is connected to a text node with degree $1$; hence, we use a red edge to indicate that the skill node $a$ is learned by the text node. Consequently, the skill node $a$ can be learned, allowing the edges connected to it to be removed from the skill-text bipartite graph. \rfig{SCNS}(c) shows the graph after the edges are removed, revealing that the skill node $b$ is connected to a text node with degree $1$, indicating it can also be learned. \rfig{SCNS}(d) represents the last iteration, where no additional skills can be learned, concluding the process with the skills $a$ and $b$ being learned, while skills $c$ and $d$ are not.

This SCNS training process is mathematically equivalent to the iterative decoding approach in LDPC codes over the binary erasure channel (BEC) \cite{gallager1962low,shokrollahi1999new,richardson2001design} and Irregular Repetition Slotted ALOHA (IRSA) \cite{liva2011graph,narayanan2012iterative,paolini2012random,jakovetic2015cooperative,stefanovic2018coded,chiang2022parallel}. Following the density evolution analysis (in the asymptotic regime of
a large number of skills) \cite{luby1998analysis,luby1998analysisb,richardson2001capacity}, {\yw we can derive the probability
that a randomly selected text is not understood by the learner. This can be viewed as the {\cs training} error of the learner as a result of the above training process. In particular,} let $q^{(i)}$
be the probability that the skill end
of a randomly selected edge is not learned after  the $i^{th}$ {\cs SCNS} iteration of training. Initially, we have $q^{(0)} = 1$ (as all the skills are novel before training). After the $i^{th}$ {\cs SCNS} iteration, the number of novel skills in a randomly selected text follows a Poisson distribution with mean $q^{(i)} c$
(a more detailed argument for this is provided in \rsec{appmul}).
For the 1-skill learner, a novel skill can be learned if it is the only novel skill in a text, which happens with probability $e^{-q^{(i)}c}$. Thus, the probability that the text end of a randomly selected edge does not contribute to the learning of
a novel skill in the $(i+1)^{th}$ SCNS iteration is
\beq{one1111}
p^{(i+1)} = 1 - e^{-q^{(i)} c}.
\eeq

To compute $q^{(i+1)}$, let us consider a randomly selected edge. Suppose that the  skill (resp. text) end  of that edge is $s$ (resp. $t$).
By SCNS, the skill $s$ is not learned
 after the $(i+1)^{th}$  SCNS iteration if the text nodes connected to $s$ (excluding the text node $t$ from the randomly selected edge) are not able to contribute to the learning of skill $s$ in
%were not able to learn the skill $s$ after
the $(i+1)^{th}$ SCNS iteration.  Since the excess degree of a Poisson degree distribution is still Poisson with the same mean \cite{Newman2010}, the number of text nodes connected to $s$ (excluding the text node $t$) is a Poisson random variable with mean $cR$ (from \rprop{skilldegree}(i)). Thus,
\beq{one2222}
q^{(i+1)} = \sum_{k=0}^{\infty} e^{-cR}\frac{(cR)^k}{k!} {p^{(i+1)}}^k = e^{-c R (1-p^{(i+1)})}.
\eeq

Letting $i \to \infty$ in \req{one1111} and \req{one2222} yields
\beq{one3333}
p = 1 - e^{- c e^{-c R(1-p)} }.
\eeq
Here, $p$ represents the probability that the text end of a randomly selected edge still contains more than one novel skill at the end of the training process (i.e., after the learner has been repeatedly presented with the set of training texts $\calD$).
Equation \req{one3333} characterizes the scaling law of the value $p$ with respect to $R$, the ratio of the number of training texts to the number of skills.

\begin{figure}[t]
    \centering
    \includegraphics[width=0.8\columnwidth]{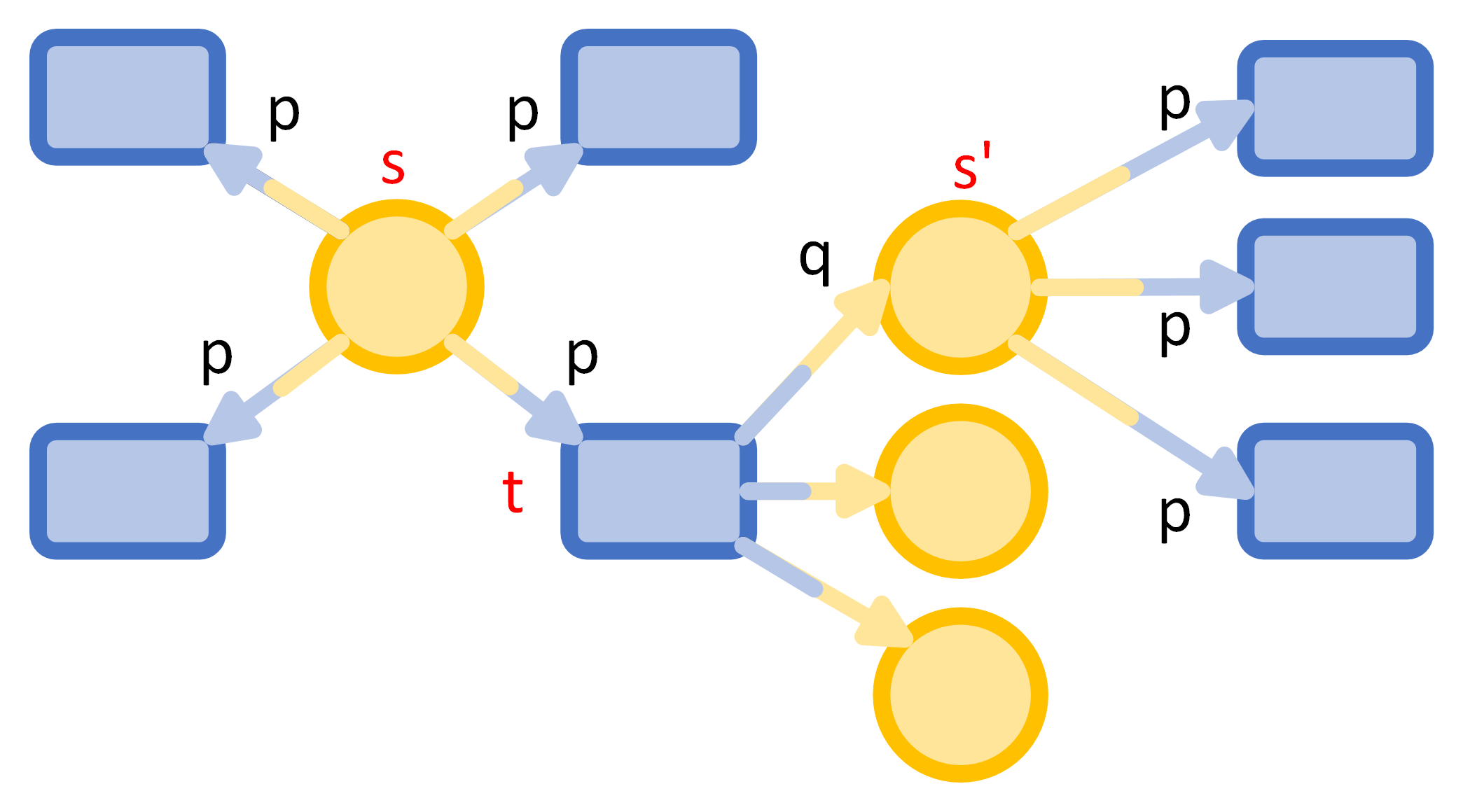}
   \caption{We illustrate the density evolution (tree) analysis by considering a skill node $s$. The skill node $s$  remains novel if it cannot be learned from any text connected to $s$. This leads to \req{zata1111} (as the degree of node $s$ is Poisson with mean $cR$).
     For the 1-skill learner, the text end $t$ of an edge connected to $s$ can learn $s$ if it is the only edge connected to $t$, i.e., all the other edges connected to $t$ (excluding the edge between $s$ and $t$) are removed. This leads to \req{one1111} (as the excess degree of node $t$, excluding the edge between $s$ and $t$, is Poisson with mean $c$and each excess edge is removed with probability $q$). Now let us go down the tree one level further. The edge between $t$ and $s^\prime$ cannot be removed if $s^\prime$ is not learned (remains novel). The skill node $s^\prime$ remains novel if it cannot be learned from any text connected to $s^\prime$ (excluding the text $t$). This leads to \req{one2222} (as the excess degree of node $s^\prime$, excluding the edge between $s^\prime$ and $t$, is Poisson with mean $cR$).}	
    \label{fig:skill}
   \end{figure}

Now a skill $s$ remains novel after training if it cannot be learned from any text connected to $s$.
(see \rfig{skill} for an illustration).
As the degree of a skill node is Poisson distributed with mean $c R$, this occurs with probability $e^{-cR (1-p)}$. Thus, the probability that a randomly selected skill is learned is
\beq{zata1111}
\zeta = 1 - e^{-cR (1-p)}.
\eeq

\begin{figure}[t]
    \centering
    \includegraphics[width=0.75\columnwidth]{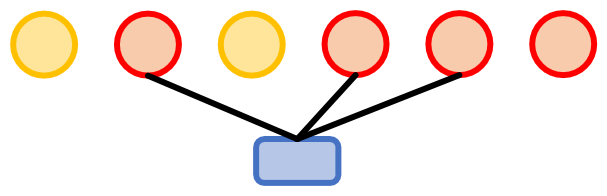}
    \caption{{\cs A randomly selected text can be understood if all the connected skill nodes are learned (highlighted in red circles). This leads to \req{one4444}.}}	
    \label{fig:test}
    \end{figure}

 For testing, we randomly select a text $t$ from $\calT$ (not included in the set of training texts $\calD$).
  {\cs Assume that this randomly selected text has the same distribution as those in the training data, i.e., it also satisfies (A1) and (A2).
  As such, the number of skills in $t$ is also Poisson distributed with mean $c$. Recall that a text can be understood if all the connected skill nodes are learned (see \rfig{test} for an illustration). Define the testing error (denoted by $\epsilon$) as the probability that the randomly selected text is not understood by the learner. For the 1-skill learner, we have
\bear{one4444}
\epsilon&=&1-\sum_{k=0}^{\infty} e^{-c} \frac{c^k}{k!} \zeta^k \nonumber\\
&=&1- e^{-c(1-\zeta)} = 1-e^{-c e^{-c R(1-p)}}.
\eear
In this case, the probability of the testing error is also equal to the solution for $p$ in \req{one3333}, for $0 \le p \le 1$.
Since $p \ge 0$, we know from \req{one4444} that the testing error is at least $1-e^{-c e^{-c R}}$.}

%\iffalse
%\fi

%In view of \req{one3333}, the testing error for the 1-skill learner is $1-p$, where $0 \le p \le 1$ is the largest solution of \req{one3333}. Since $p \ge 0$, we know from \req{one4444} that the testing error is at least $e^{-c e^{-c R}}$.

\begin{figure}[t]
	\centering
	\includegraphics[width=0.48\textwidth]{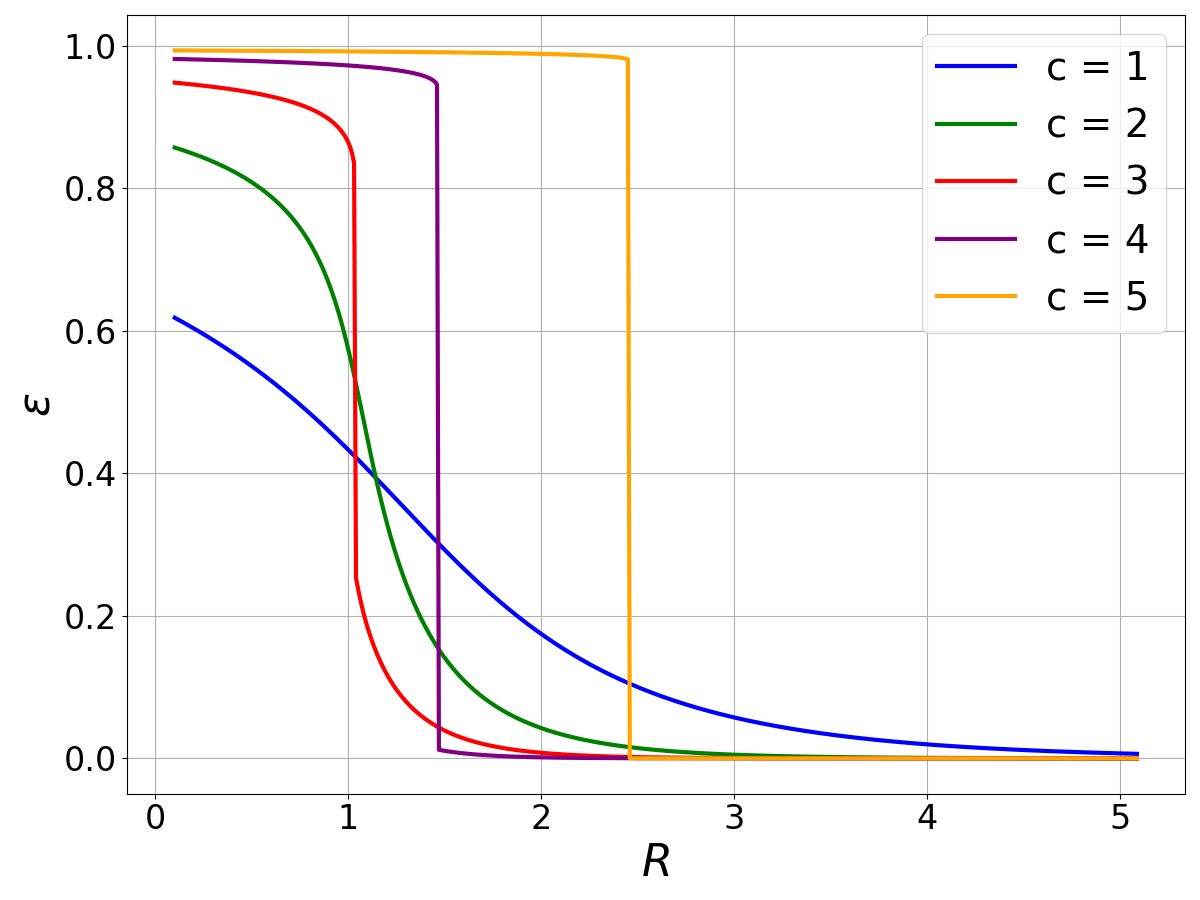}
	\caption{The plot illustrates the probability of the testing error \( \epsilon \) as a function of \( R \) for the 1-skill learner.}
%illustrates the computed values of \( \epsilon \) as a function of \( R \), which ranges from 0.1 to 5 with a step size of 0.1, for each \( c \) value (1, 2, 3, 4, 5). The computation is based on the recursive equation \( p^{(i+1)} = 1 - e^{-( c e^{-c R(1-p^{(i)})} )} \) with the initial condition \( p^{(0)} = 1 \).}
	\label{fig:pvalues}
\end{figure}

In \rfig{pvalues}, we illustrate the effect of the parameters $R$ and $c$ on the probability of the testing error $\epsilon$. We compute $p$ by using the recursive equation $p^{(i+1)} = 1 - e^{- c e^{-c R(1-p^{(i)})}}$ with the initial condition $p^{(0)} = 1$. The number of iterations is set to 10,000.
We can see that there exists a percolation threshold, indicated by a sudden change in the values of $\epsilon$. The phase transition behavior is most evident when the average number of skills contained in each text (i.e., $c$) is greater than $3$. For $c = 3$ (respectively, 4 and 5), this threshold occurs when $R$ exceeds approximately 1 (respectively, 1.5 and 2.5).
The underlying rationale is that for $c \ge 3$, a randomly chosen text is unlikely to encompass precisely one novel skill. Consequently,
a substantial volume of training texts is essential to ensure a sufficient quantity of texts initially containing only a single novel skill. This foundation of learned skills subsequently initiates a cascade of successive cancellations of novel skills.

\bsubsec{Poisson Learner}{Poisson}

In this section, we extend the concept of the 1-skill learner to the Poisson learner, where multiple novel skills may be learned from a text with a certain probability.

\bdefin{Poisson} ({\bf Poisson Learner}) An abstract learner is termed a {\em Poisson learner} if a novel skill $s$ in a text $t$ can be learned with probability $P_{\rm suc}(\rho)$ when the number of {\em novel} skills in that text is Poisson distributed with mean $\rho$. Once a skill $s$ is learned, the learner is able to identify
the skill $s$ appearing in any other text.
\edefin

The definition of Poisson learners is analogous to that of Poisson receivers in \cite{chang2020Poisson,liu2021aloha,chang2022stability} for iterative decoders.
As discussed in \cite{chang2020Poisson,liu2021aloha,chang2022stability}, a 1-skill learner is a Poisson learner with the success probability $P_{\rm suc}(\rho)=e^{-\rho}$. Similarly, the 2-skill learner, where skills in a text can be learned if the number of novel skills does not exceed two, is a Poisson learner with the success probability $P_{\rm suc}(\rho)=e^{-\rho} + \rho e^{-\rho}$ \cite{chang2020Poisson,liu2021aloha,chang2022stability}. Using the density evolution method from \cite{luby1998analysis,luby1998analysisb,richardson2001capacity}, it was shown in \cite{chang2020Poisson,liu2021aloha,chang2022stability} that
\bear{tag6677}
p^{(i+1)}&=& 1-P_{\rm suc}(q^{(i)} c), \label{eq:tag6677a}\\
q^{(i+1)}&=&e^{-c R(1-p^{(i+1)})} \label{eq:tag6677b}.
\eear
Letting $i \to \infty$ yields
\beq{one6666}
p = 1 - P_{\rm suc}(c e^{-c R(1-p)}).
\eeq
Then, analogous to the argument for the 1-skill learner, the testing error for a Poisson learner is characterized in \req{one4444} with $p$ given
in \req{one6666}.

\begin{figure}[t]
	\centering
	\includegraphics[width=0.48\textwidth]{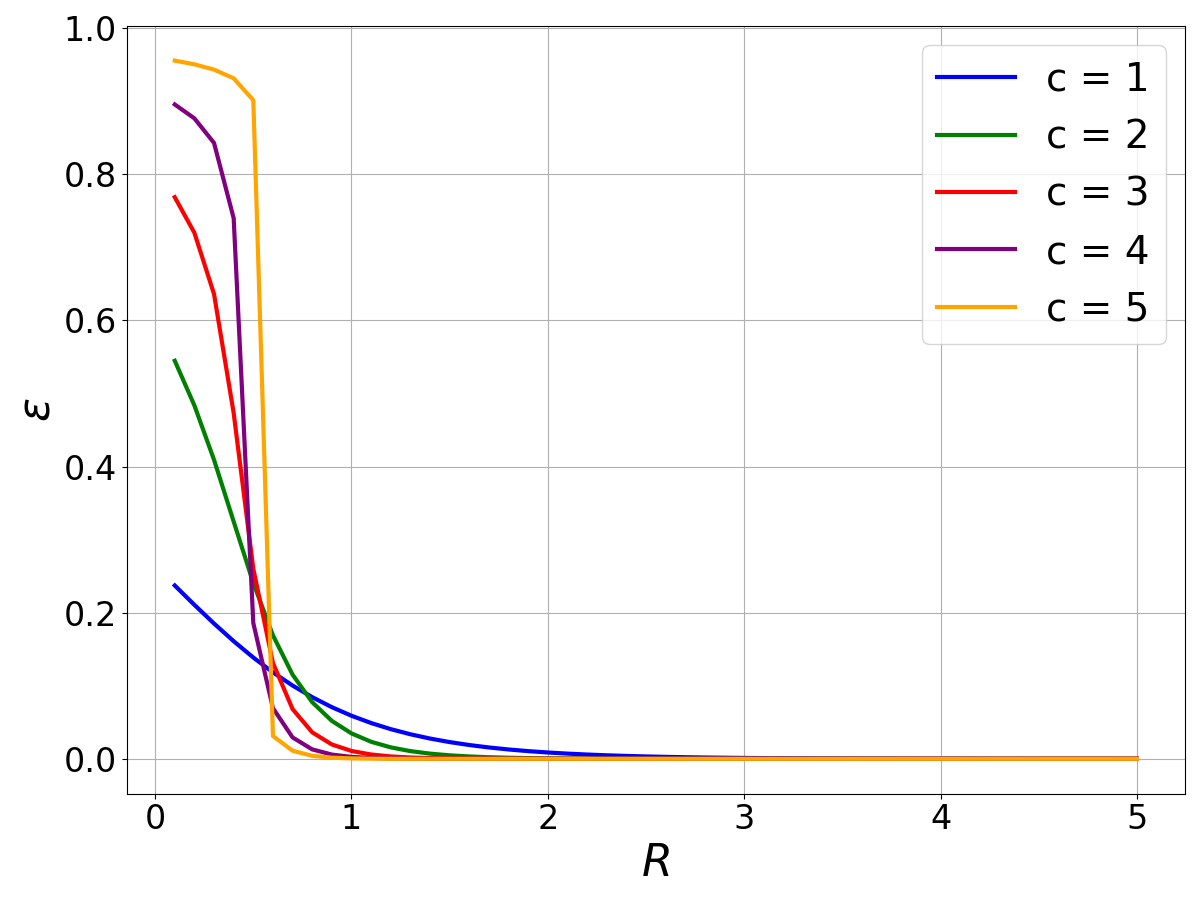}
	\caption{The plot illustrates the probability of the testing error \( \epsilon \) as a function of \( R \) for the 2-skill learner.}
%, which ranges from 0.1 to 3 with a step size of 0.1, for each \( c \) value (1, 2, 3, 4, 5). The computation is based on the recursive equation \( p^{(i+1)} = 1 - P_{\rm suc}( c e^{-c R(1-p^{(i)})} ) \) with the initial condition \( p^{(0)} = 1 \), where $P_{\rm suc}(\rho)=e^{-\rho}+\rho e^{-\rho}$.}
	\label{fig:p2values}
\end{figure}

In \rfig{p2values}, we illustrate the effect of the parameters $R$ and $c$ on the probability of the testing error $\epsilon$ for the 2-skill learner. We first compute $p$ by using
the recursive equation $p^{(i+1)}=1-P_{\rm suc}( c e^{-c R(1-{p^{(i)}})} ) $ with the initial condition \( p^{(0)} = 1 \) and $P_{\rm suc}(\rho)=e^{-\rho}+\rho e^{-\rho}$. The number of iterations is set at 10,000. A percolation threshold, indicated by a sudden change in the values of $\epsilon$, exists for $c = 4$ (respectively, 5) when $R$ is approximately 0.5 (resp. 0.6). The percolation thresholds for the 2-skill learner are significantly lower than those for the 1-skill learner, demonstrating that the 2-skill learner is a much faster learner than the 1-skill learner.

\iffalse % red
{\red [I am curious how the test error curves look in log-scale. Can there be any insights gained on the decay rate of testing error.]} {\cs This is an interesting point as we may be able to obtain the scaling law by looking the error curves in log-scale. But I am not sure if we got time for this before the deadline of the submission.}
\fi

\bsec{Learning the association of skills}{skillclusters}

When the training is complete, a learner can learn not only a fraction of skills but also the associations between these learned skills. We say two learned skills are {\em associated} if they appear in the same text. {\yw Since a learned skill can be identified whenever it is present in a text,}
we can generate a skill association graph  for learned skills by adding an edge between two learned skill nodes if they appear in the same text.
For instance, the two learned skills, $a$ and $b$, shown in \rfig{SCNS}, are associated as they appear in the same text. After training, we can generate a skill association graph with nodes $a$ and $b$, and an edge connecting these two nodes.

Knowing the structure of the skill association graph is crucial, as it can be utilized for inference purposes, such as predicting the next skill given a set of skills. This is analogous to giving an LLM a {\em prompt} in the form of a text with a set of skills and asking the LLM to predict the next skill and generate a text based on the predicted skill. An essential question arises: Are most of the learned skills interconnected in a way that facilitates their use for prediction? This question guides us to perform a giant component analysis of the skill association graph.
Note that a giant component in a random graph is a connected subgraph whose size is proportional to the size of the graph. It is well-known  \cite{Newman2010} that there exists at most one giant component in a random graph.
The rest of the components are called {\em small} components. One important property of a random graph is that small components are {\em trees} with high probability.

Our approach to this problem is based on the site (or node) percolation analysis as described in \cite{Newman2010}.
As in \rsec{Poisson}, we consider the skill-text bipartite graph with the set of skill nodes $\calS$ and the set of text nodes $\calD$.
It is important to note that not every skill is learned upon the completion of training. In fact, as shown in \req{zata1111}, the probability that a randomly selected skill node is learned is given by
\beq{ps1111}
\zeta=1 - e^{-cR (1-p)},
\eeq
where $p$ is the solution of \req{one6666}.

To apply the  site percolation analysis for the skill-text bipartite graph, we assume that skill nodes are learned {\em independently} with probability $\zeta$.
%Let $\mu_s$ (respectively, $\mu_t$) be the probability that a randomly selected edge of a skill node (respectively, text node) is only connected to a small component.
%  via one of  its edges}.
Let $\mu_s$ be the probability that a skill node is connected to a small component via one of its edges.
Also, let $\mu_t$ be the probability that a text node is connected to a small component via one of  its edges.
Recall from \rprop{skilldegree} that  the degree distribution of the text nodes (respectively, skill nodes) is Poisson with mean $c$ (respectively, $c R$).
%{\red Note that a randomly selected edge (called the tagged edge) of a skill node is only connected to a small component}
Note that a skill node is only connected to a small component via one of  its edges (called the tagged edge)
if all the excess edges of the text end of the tagged edge are only connected to small components.
Since the excess degree distribution of a Poisson degree distribution is also Poisson with the same mean, we then have (from the tree property for small components) that
\beq{sclu11110}
\mu_s=\sum_{k=0}^\infty e^{-c} \frac{c^k}{k!} \mu_t^k
= e^{-c(1-\mu_t)}.
\eeq
Now we derive the governing equation for $\mu_t$.  If the skill end of an edge (called the tagged edge) is not learned,
then there are no further connected skill nodes via this edge, which occurs with probability $1-\zeta$. Conversely, if the skill end of the tagged edge is learned, then it is only connected to a small component if all the excess edges of the skill end of the tagged edge are only connected to small components. Thus, we have (from the tree property for small components) that
\bear{sclu22220}
\mu_t&=&(1-\zeta)+\zeta\sum_{k=0}^\infty e^{-cR} \frac{{(cR)}^k}{k!} \mu_s^k \nonumber\\
&=& (1-\zeta) +\zeta e^{-Rc(1-\mu_s)}.
\eear
Clearly, $\mu_s=\mu_t=1$ is a trivial solution of \req{sclu11110} and \req{sclu22220}. To find a nontrivial solution (if any), one can iterate
the following two equations:
\bear{sclu33330}
&&\mu_s^{(i+1)}=e^{-c(1-\mu_t^{(i)})}, \nonumber\\
&&\mu_t^{(i+1)}=(1-\zeta)+ \zeta
 e^{-Rc(1-\mu_s^{(i)})},
\eear
with the initial condition $\mu_s^{(0)}=\mu_t^{(0)}=0$. {\cs By induction, it is easy to see from \req{sclu33330} that $0 \le \mu_s^{(i)} < \mu_s^{(i+1)} \le 1$ and $0 \le \mu_t^{(i)} < \mu_t^{(i+1)} \le 1$ for all $i$. Thus, the recursion in \req{sclu33330} converges to a solution in $[0,1] \times [0,1]$ as $i \to \infty$.}

Note that a randomly selected skill node is in the giant component if the skill node is learned and {\cs at least one of its edges is not connected to a small component}. Thus,
the probability that a randomly selected skill node is in the giant component, denote by $p_G$, is
\beq{sclu4444}
p_G={\zeta}\Big(1- \sum_{k=0}^\infty e^{-cR} \frac{{(cR)}^k}{k!} \mu_s^k\Big) =\zeta(1-e^{-Rc(1-\mu_s)}).
\eeq

\begin{figure}[t]
	\centering
	\includegraphics[width=0.48\textwidth]{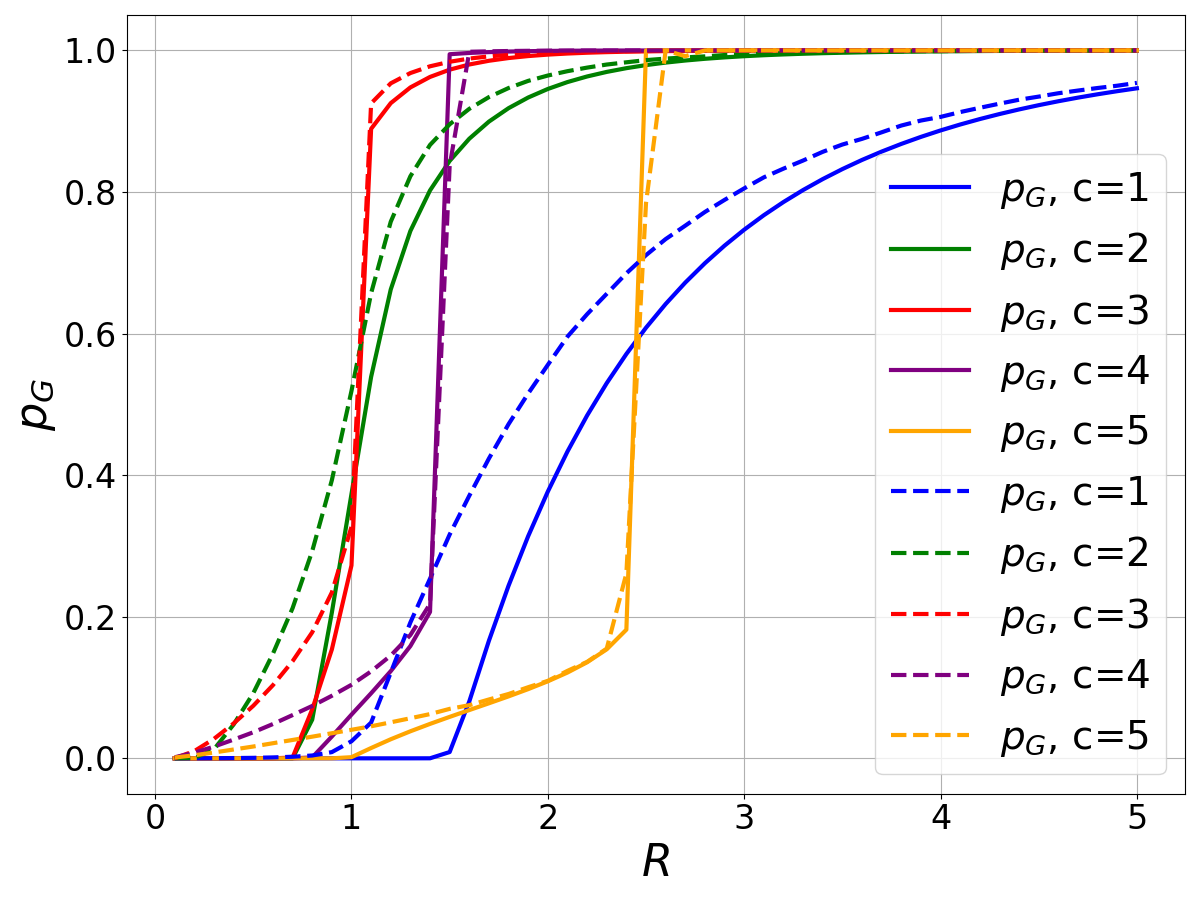}
	\caption{The plot of the probability $p_G$ that a randomly selected skill is in the giant component for the 1-skill learner as a function of $R$ for various values of $c$. }
	\label{fig:giantsite2}
\end{figure}

In \rfig{giantsite2}, we present a plot of the probability $p_G$ that a randomly selected skill is in the giant component as a function of $R$ for the 1-skill learner with values of $c=1,2,3,4,5$. The solid line represents the theoretical value, and the dashed line represents the simulated value obtained by averaging over 100 random bipartite graphs with a total of 20,000 skill and text nodes. This figure offers an intriguing comparison to the testing errors shown in \rfig{pvalues}. For $c=1$, as indicated in \rfig{pvalues}, the testing errors decrease with an increase in $R$. However, the learned skills do not form a giant component until $R$ surpasses a threshold, approximately at 1.5, as depicted in \rfig{giantsite2}. While it is relatively easy for the 1-skill learner to acquire skills when the average number of skills in a sampled text is small, it is also challenging to learn the association between two learned skills. Conversely, for $c=5$, learning individual skills is difficult. Yet, when $R$ exceeds a threshold, roughly at 2.5, we observe a significant reduction in testing errors in \rfig{pvalues}. Additionally, surpassing this same threshold in \rfig{giantsite2} correlates with the emergence of a giant component with the size near the size of the skill nodes. This suggests that the acquisition of skills and their associations occur simultaneously once this threshold is exceeded.  It is noteworthy that discrepancies exist between the theoretical and simulation results, as seen in \rfig{giantsite2} for $c=1$ and $2$. These discrepancies arise because the percolation analysis assumes skill nodes are learned independently--a premise that does not hold when
$c$ is small.

Next, we derive the condition for the existence of a giant component.
Combining \req{sclu11110} and \req{sclu22220} yields
{\cs
\bear{condition1111}
\mu_s &=&  e^{-c\{1 - [ (1-\zeta) +\zeta e^{-Rc(1-\mu_s)}] \}} \nonumber\\
&=& e^{ -c \zeta ( 1 - e^{-Rc(1-\mu_s)}) }.
\eear
}
{\yw Consider the function
\beq{condition2222}
g_1( \mu_s ) = e^{ -c \zeta ( 1 - e^{-Rc(1-\mu_s)}) }.
\eeq
Note that $g_1( \mu_s )$ is a convex function of $\mu_s$ when $0 \le \mu_s \le 1$. This can be verified by showing that the second derivative
of $g_1( \mu_s )$ is nonnegative for $0 \le \mu_s \le 1$. Clearly, $\mu_s = 1$ is the trivial solution of \req{condition1111}.
By the convexity and the fact that $g_1(0) > 0$, we have a non-trivial solution for some $\mu_s < 1$ if the slope of $g_1(\mu_s)$ at $\mu_s = 1$ is greater than the slope of the line $y=\mu_s$, i.e.,
\beq{condition3333}
\left.{\frac{d}{d\mu_s}g_1( \mu_s )} \right|_{\mu_s = 1}  > 1.
\eeq
This condition is equivalent to
%A straightforward algebra shows that the condition in \req{condition3333} is equivalent to
\beq{condition4444}
c^2R\zeta  > 1
\eeq
or, by the representation of $\zeta$ in \req{ps1111},
\beq{condition4444b}
c^2R(1-e^{-cR(1-p)}) > 1.
\eeq
Here, $p$ is the probability that the text end of a randomly selected
edge is not learned and is given by the solutions of \req{one3333} (respectively, \req{one6666}) for a 1-skill learner (respectively, Poisson learner).
In \rfig{mus}, we provide an illustration of the condition in \req{condition3333}  for $c=3$ and $R=0.5, 1, 1.5$ by considering a 1-skill learner. We can see that a non-trivial solution exists at the intersection between the curve $y = g_1(\mu_s)$ and the line $y = \mu_s$ for all these cases.

\begin{figure}[t]
	\centering
	\includegraphics[width=0.48\textwidth]{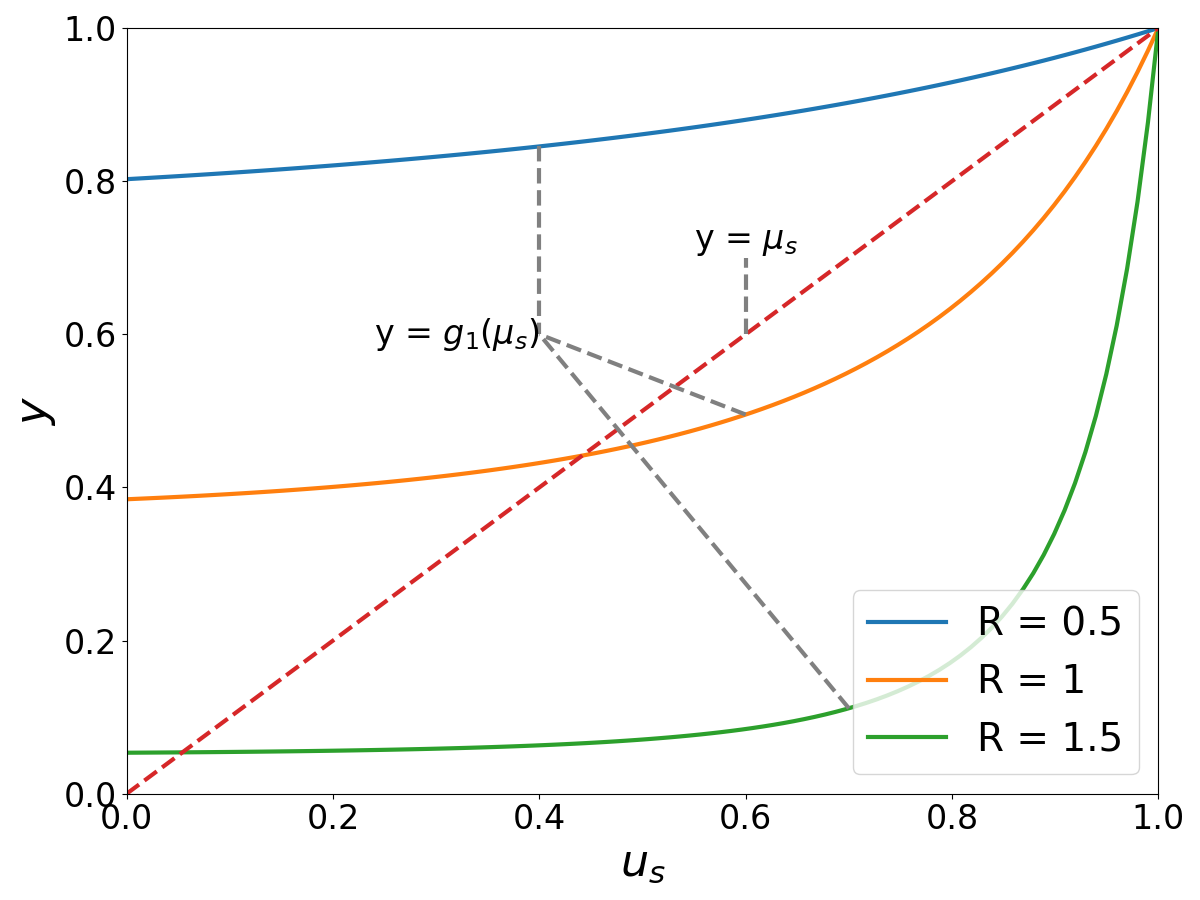}
	\caption{The solution of the equation \req{condition1111} is found at the point where the curve $y = g_1(\mu_s)$ intercepts the line $y = \mu_s$, given that $c = 3$. }
	\label{fig:mus}
\end{figure}

\begin{figure}[t]
	\centering
	\includegraphics[width=0.48\textwidth]{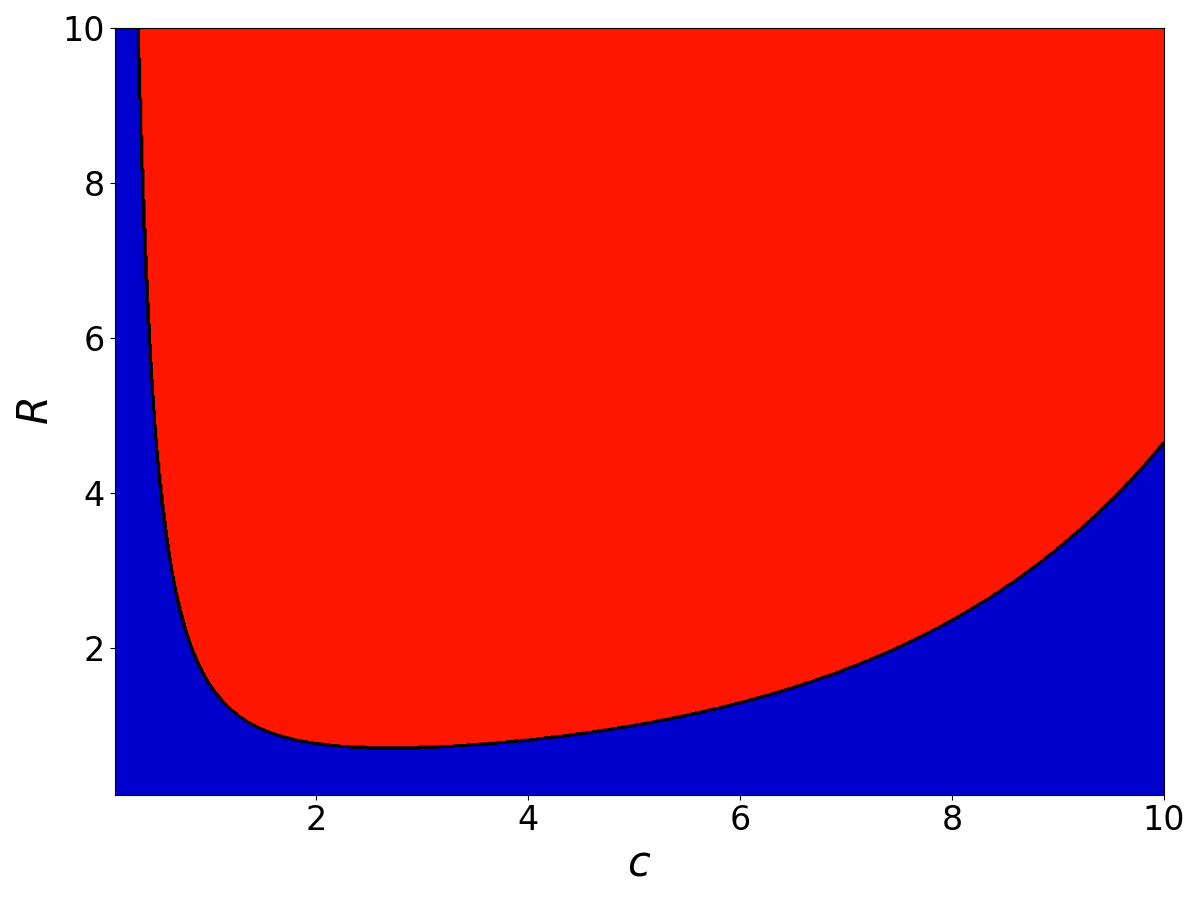}
	\caption{Regions based on the slope of the function $g_1( \mu_s )$ at
$\mu_s=1$. The red region corresponds to $g_1( \mu_s )>1$, indicating the existence of a giant component, and the blue region corresponds to $g_1( \mu_s )<1$, signifying the absence of a giant component.}
	\label{fig:ccRps}
\end{figure}

For each pair of $c$ and $R$, we can check whether the condition in  \req{condition4444b} is satisfied. If it is satisfied, then {\yw there exists a non-trivial solution for $\mu_s$, implying the existence of a giant component}. In \rfig{ccRps}, we show the region (marked in red) where a giant component emerges. On the other hand, the region (marked in blue) does not exist a giant component.
As shown in \rfig{ccRps}, it is observed that \(R\) must be sufficiently large for both scenarios where \(c\) is large and \(c\) is small. Specifically, when \(c\) is large (e.g., \(c>2\)), a large \(R\) is necessary to surpass the percolation threshold for \(p\) in \req{one3333}. Surpassing this threshold leads to the learning of a large number of skills, thereby facilitating the automatic emergence of a giant component within the skill association graph due to a large number of skills in a text. Conversely, when \(c\) is small (e.g., \(c<1\)), nearly every skill in a text is learned by the 1-skill learner (since the average number of skills per text is less than 1). This results in a very small \(p\). Nonetheless, even in this scenario, a large \(R\) is required to learn a sufficient number of skills to meet the threshold for a giant component's existence as stipulated in \req{condition4444b}.
We also note from \rfig{ccRps} that the minimum $R$ to have a giant component is 1.1 and this is achieved when  $c = 1.29$.
%This implies that if the training text contains an average of $1.29$ skills, then the minimum number of texts required to form a giant component is $1.1$.

\bsec{Hierarchy of skills}{hskills}

One common approach to train a domain-specific LLM is to adopt a pre-trained model, commonly referred to as a foundation model or a basic model, and fine tune it with additional domain-specific texts.
Motivated by this, we
consider the setting where there are two classes of skills: {\yw the class of basic skills $\calS$ and the class of domain-specific skills $\calS_f$ (see \rfig{hier} for an illustration). The original set of texts $\calT$ is treated as a basic set of texts, which requires only basic skills to understand. In addition, we also consider an additional set of domain-specific texts $\calT_f$, {\cs where each requires learning a certain number of domain-specific skills to understand.
A prerequisite of learning a domain-specific skill requires learning a random number of basic skills first.
Thus, there is a hierarchy of skills.}
%These domain-specific skills (resp. texts) can only be learned (resp. understood) after {\cs  basic skills are successfully learned, thus forming a hierarchy of skills.

\begin{figure}[t]
	\centering
	\includegraphics[width=0.48\textwidth]{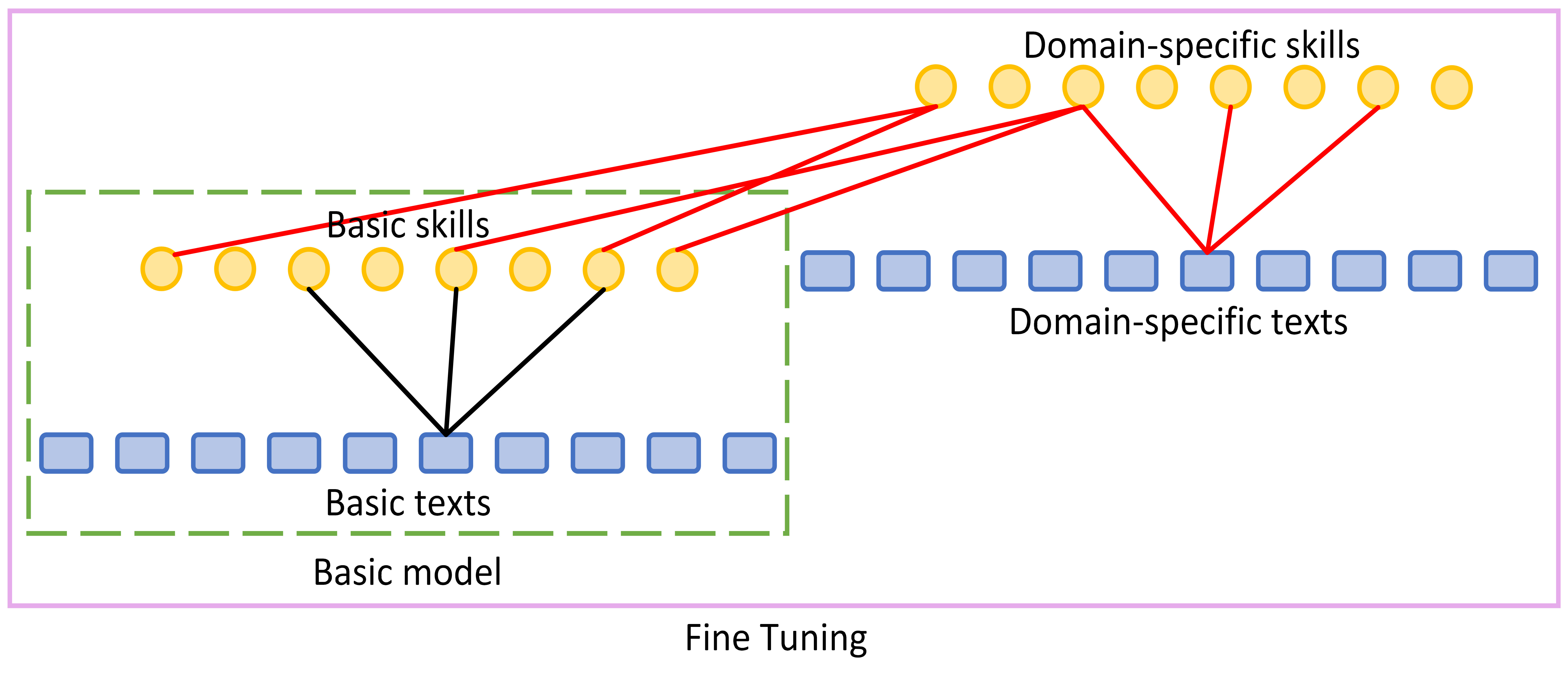}
	\caption{This figure is an illustration of the hierarchy of skills. The basic model is trained first and then fine-tuned with domain-specific texts.  A prerequisite of learning a domain-specific skill requires learning a random number of basic skills.}
%{\red [I would suggest to draw the domain-specific texts on the same level as the basic texts, and also have dashed lines connecting to some basic skills. The idea is that, these domain-specific texts may also contain some basic skills, but they are decodable once the threshold for basic skills is reached.]}}
	\label{fig:hier}
\end{figure}

{\yw In particular, following the procedure in \rsec{one}, we first train a basic model (foundation model) for a 1-skill learner using a set of basic texts $\calD$ sampled from $\calT$.
Then, the probability that a randomly selected basic skill is learned can be computed as} $\zeta = 1 - e^{-cR (1-p)}$, where $p$ is the solution to \req{one3333}.
%After the basic model is trained, it is fine-tuned to obtain a domain-specific model by training on a set of domain-specific texts $\calD_f$ sampled from $\calT_f$.}
%We assume that there are $|\yw\calS_f|$ domain-specific skills, and a prerequisite of learning a domain-specific skill requires learning a random number of basic skills.
%Thus, there is a hierarchy of skills.
Suppose that the basic skills required by a  domain-specific skill are uniformly selected from the set of basic skills $\yw\calS$. Let $\delta_\ell$ be the probability {\yw that} the number of basic skills required by a  domain-specific skill
is $\ell$ and $\Lambda_f(x)=\sum_{\ell=0}^\infty \delta_\ell x^\ell$ be the corresponding generating function.
Then we say a domain-specific skill is {\em learnable} if all its basic skills are learned in the basic {\yw model. Hence, it occurs} with probability
\beq{hier1111}
\sum_{\ell=0}^\infty \delta_\ell {\zeta}^\ell=\Lambda_f(\zeta).
\eeq

{\yw To fine-tune} the basic model, we
 collect a subset of  domain-specific texts $\calD_f$ and present these $|\calD_f|$ texts to {\yw a 1-skill learner to acquire domain-specific skills.
  %(as a way of fine tuning the basic model).
Let} $R_f = |\calD_f|/|S_f|$ be the ratio of the number of training domain-specific texts to the number of domain-specific skills.
Similar to the basic model, we make the following two assumptions about sampled domain-specific texts:
\begin{description}
\item [(A1d)] (Poisson degree distribution) Denote by $N_f(t)$ the number of domain-specific skills in the domain-specific text $t \in  \calD_f$.
Then $\{N_f(t), t \in  \calD_f\}$ are independent Poisson random variables with mean $c_f$, i.e., $\pr(N_f(t) = k) = e^{-c_f} \frac{c_f^k}{k!}$ for $k = 0, 1, 2, \ldots$.
\item [(A2d)] (Uniform connection) Each edge of a domain-specific text node is connected to a domain-specific skill node independently and uniformly. Thus, the probability that an edge of the domain-specific text node $t$ is connected to a particular domain-specific skill node $s$ is $1/|\calS_f|$.
\end{description}

A domain-specific text node is said to be {\em learnable} if all {\yw of} the domain-specific skill nodes connected to it are learnable. Otherwise, it is called a {\em non-learnable} domain-specific text node. We assume that nothing can be learned from a non-learnable domain-specific text when it is presented to a learner.
From (A1d), (A2d) and \req{hier1111}, we deduce that the probability a randomly selected domain-specific text is learnable is given by
\beq{hier2222}
\sum_{\ell=0}^\infty e^{-c_f} \frac{ c_f^\ell}{\ell!} (\Lambda_f(\zeta))^\ell=e^{-c_f (1-\Lambda_f(\zeta))}.
\eeq

Define the bipartite graph formed by the learnable domain-specific skill nodes and the learnable domain-specific text nodes as the {\em learnable domain-specific bipartite graph}. In such a bipartite graph, there are approximately $\Lambda_f(\zeta) |\calS_f|$ learnable domain-specific skill nodes and $e^{-c_f (1-\Lambda_f(\zeta))} |\calD_f|$ learnable domain-specific text nodes. Furthermore, from Bayes' formula, we deduce that the degree of a randomly selected learnable domain-specific text node $t$ in the learnable domain-specific bipartite graph follows a Poisson random variable with mean $c_f \Lambda_f(\zeta)$ since
\begin{align}
&\pr (\mbox{the degree is}\; \ell | \mbox{a  domain-specific text is learnable}) \nonumber\\
&=\frac{e^{-c_f} \frac{ c_f^\ell}{\ell!} (\Lambda_f(\zeta))^\ell}{e^{-c_f (1-\Lambda_f(\zeta))}}=e^{-c_f \Lambda_f(\zeta)}\frac{ (c_f \Lambda_f(\zeta))^\ell  }{\ell!}\label{eq:hier3333}
\end{align}
This leads to the following proposition:
\bprop{hier}
Consider the learnable domain-specific bipartite graph.
Let  $\tilde \calS_f$ (resp. $\tilde \calD_f$) be the set of {\em learnable} domain-specific skills (resp. texts). The learnable domain-specific bipartite graph satisfies the following properties:
\begin{description}
\item[(P1d)] (Poisson degree distribution)
Denote by $\tilde N_f(t)$ the number of {\em learnable} domain-specific skills in the {\em learnable} domain-specific text $t \in  \tilde \calD_f$.
Then $\{\tilde N_f(t), t \in  \tilde \calD_f\}$ are independent Poisson random variables with mean $c_f \Lambda_f(\zeta)$.
\item [(P2d)] (Uniform connection) Each edge of a {\em learnable} domain-specific text node is connected to a {\em learnable} domain-specific skill node independently and uniformly. Thus, the probability that an edge of the {\em learnable} domain-specific text node $t$ is connected to a particular {\em learnable} domain-specific skill node $s$ is $1/|\tilde \calS_f|$.
\end{description}
\eprop

In light of \rprop{hier}, we can apply the same density evolution analysis used in \rsec{learners} to analyze the learnable domain-specific bipartite graph. Suppose that we still use the 1-skill learner for fine tuning.
Let $p_f$ be the probability that the text end of a randomly selected edge in the learnable domain-specific bipartite graph
is not learned after fine tuning. {\cs Note that
$|\tilde \calS_f| \approx  \Lambda_f(\zeta) |\calS_f|$  and $|\tilde \calD_f| \approx e^{-c_f (1-\Lambda_f(\zeta))} |\calD_f|$.}
Replacing $c$ with $c_f \Lambda_f(\zeta)$ and $R$ with $R_f e^{-c_f (1-\Lambda_f(\zeta))} /\Lambda_f(\zeta)$
in \req{one3333} yields
\beq{one3333h}
p_f = 1 - \exp\Big ({- c_f \Lambda_f(\zeta) e^{-(1-p_f) c_f  R_f  e^{-c_f (1-\Lambda_f(\zeta))} } }\Big).
\eeq
Similarly, let $\zeta_f$ be the probability that a randomly selected domain-specific skill node in the learnable domain-specific bipartite graph
is  learned after fine tuning. Then (cf. \req{zata1111})
\beq{one3333g}
\zeta_f=1-e^{-(1-p_f) c_f  R_f  e^{-c_f (1-\Lambda_f(\zeta))} }.
\eeq

Now we derive the probability of the testing error of a randomly selected text from the domain-specific texts.
  Denote such a probability as $\epsilon_f$. {\cs Assume that this randomly selected domain-specific text has the same distribution as the training data, i.e., it satisfies (A1d) and (A2d).}
The testing error consists of two cases: (i) a randomly selected text from the {\yw set of domain-specific texts $\calT_f$} is {\em not learnable},
and (ii) a randomly selected text from the {\yw set of domain-specific texts $\calT_f$} is {\em learnable} but one of the domain-specific skills in that text is not learned.
As shown in \req{hier2222}, the probability that a randomly selected text from the domain-specific texts is not {\em learnable} is
$1-e^{-c_f (1-\Lambda_f(\zeta))}$.  Since the degree distribution of a domain-specific text in the learnable domain-specific bipartite graph is
Poisson with mean $c_f \Lambda_f(\zeta)$ in \req{hier3333}, the probability that one of the domain-specific skills in a learnable domain-specific  text is not learned is (cf. \req{one4444})
\bear{rep2222h}
&&1-\sum^{\infty}_{k = 0} e^{ -c_f \Lambda_f (\zeta)} \frac{ ( c_f \Lambda_f (\zeta))^ k }{k!} (\zeta_f)^k  \nonumber\\
&&= 1-e^{ -c_f \Lambda_f(\zeta) (1-\zeta_f) }.
\eear
Thus, the probability for Case (ii) is
\beq{rep2222i} (1-e^{ -c_f \Lambda_f(\zeta) (1-\zeta_f) })e^{-c_f (1-\Lambda_f(\zeta))}.
\eeq
Combining Cases (i) and (ii) in \req{rep2222h} and \req{rep2222i} leads to
\bear{one4444h}
\epsilon_f& = &1 - e^{- c_f ( 1 - \Lambda_f(\zeta) ) } \nonumber\\
&& +(1-e^{ -c_f \Lambda_f(\zeta) (1-\zeta_f) })e^{-c_f (1-\Lambda_f(\zeta))}.
\eear

\begin{figure}[t]
\centering
	\includegraphics[width=0.48\textwidth]{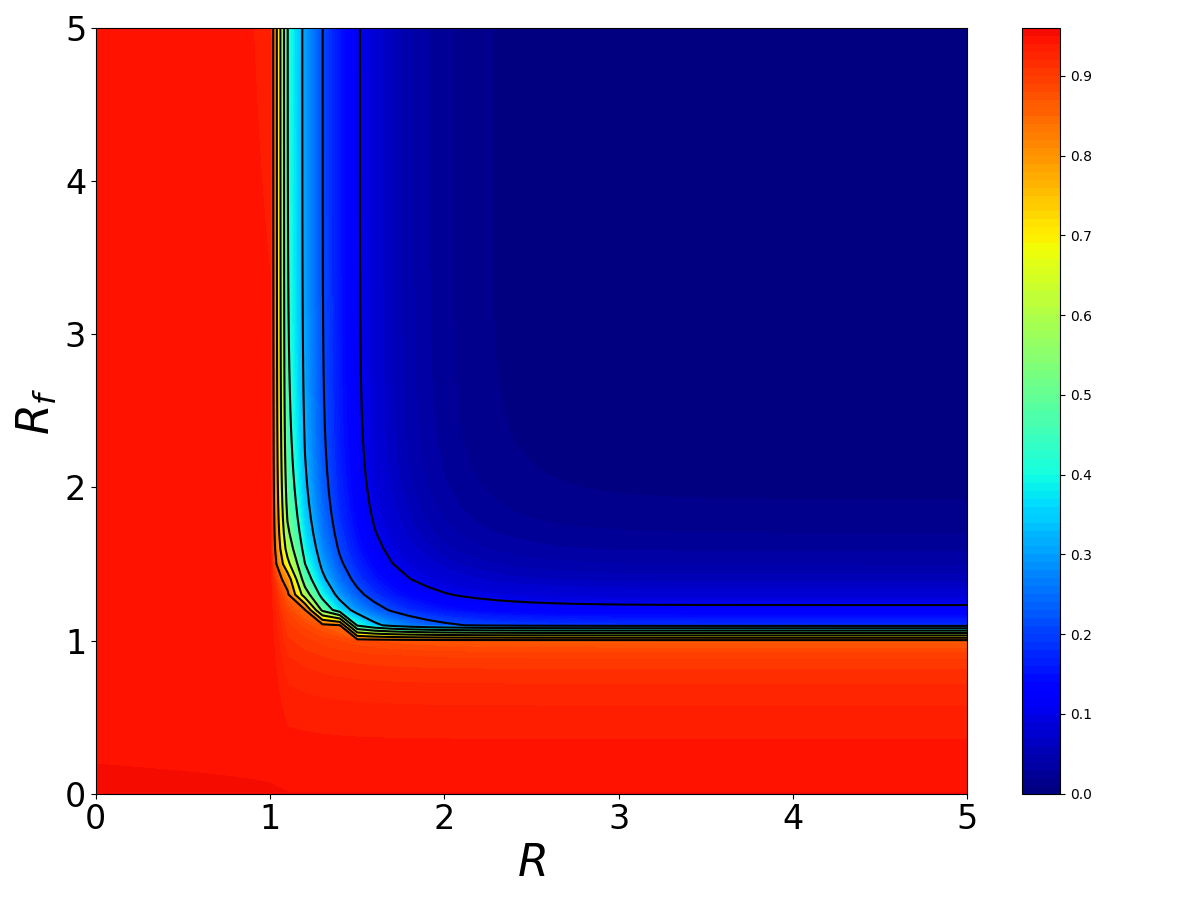}
	\caption{The contour plot of the probability of the testing error $\epsilon_f$ for the 1-skill learner as a function of $R$ and $R_f$ with values $c=3$ and $c_f = 3$. }
	\label{fig:p_t_value}
\end{figure}

In view of \req{one4444h}, there are two thresholds for achieving low testing errors in a domain-specific task: (i) a large number of basic skills are learned when the number of basic texts exceeds the threshold in the foundation model, and (ii) a large number of domain-specific skills are learned when the number of domain-specific texts exceeds the threshold in the fine-tuning model.
In \rfig{p_t_value}, we present a  contour plot of the {\yw testing error $\epsilon_f$} as a function of $R$ and $R_f$ for the 1-skill learner with values of $c = 3$ and $c_f = 3$. The number of basic skills required by a domain-specific skill is Poisson distributed with mean $3$, in which case $\Lambda_f(x) = e^{-3(1-x)}$.
The graph features two percolation thresholds associated with $R$ and $R_f$. It is observed that if $R < 1$, the majority of domain-specific text nodes remain unlearned, primarily because the basic skills are not acquired, indicating the first percolation threshold. Conversely, when $R > 1$, attention shifts to the second percolation threshold, $R_f$; in this scenario, a significant number of domain-specific texts are learned, provided that $R_f > 1$. This distinction underscores the critical values of $R$ and $R_f$ in determining the learning outcomes within the graph.

\bsec{Extensions to multiple classes of skills and texts}{ES}

In this section, we extend the analysis to the setting with multiple classes of skills and texts, which is motivated by the existence of multiple subjects in texts, such as math, physics, chemistry, law, etc. {\yw Note that, in these settings, a text may require skills from multiple classes to understand and, thus, introduces interdependence among classes. However, different from \rsec{hskills}, the multiple classes of skills are not assumed to be hierarchical in nature and, thus, could be learned without a prerequisite on the knowledge of other classes of skills.}

\bsubsec{Poisson learners with multiple classes of skills/texts}{appmul}

We  first extend the density evolution analysis for Poisson learners to settings with multiple classes of skills and texts.  Our analysis is similar to the framework of Poisson receivers in \cite{chang2022stability}.

\subsubsection{The ensemble of random bipartite graphs}
\label{sec:ensemble}

Suppose that the set of texts $\yw \calT$ can be classified into $J$ disjoint classes of texts, $\yw \calT_1, \yw \calT_2, \ldots, \yw \calT_J$,
and that the set of skills $\calS$  can be classified into $K$ disjoint classes of skills, $\yw \calS_1, \yw \calS_2, \ldots, \yw \calS_K$.
As before, we sample a subset of  texts $\calD$ from the set of texts $\yw \calT$ and present these $|\calD|$ texts as the {\em training texts}. Let $R =|\calD|/|\yw \calS|$ be the ratio of the number of training texts to the number of skills. Let $\calD_j$ be the set of class $j$ texts in $\calD$ and
$\alpha_j=|\calD_j|/|\calD|$ be the ratio of the number of class $j$ training texts to the total number of training texts, $j=1,2, \ldots, J$.
Also, let $\beta_k=|\yw \calS_k|/|\calS|$, $k=1,2, \ldots, K$, be the ratio of the number of class $k$ skills to the total number of skills.
As before,  the skill-text bipartite graph of a {\em sampled} semantic language is a random bipartite graph with $|\yw \calS|$ skill nodes on one side and $|\calD|$ text nodes on the other side. An edge exists between a skill node $s$ and a text node $t$ if $s \in \phi(t)$. {\yw An edge is called a class $(k,j)$-edge if it connects a class $k$ skill node to a class $j$ text node.}

We make the following two assumptions about the sampled texts:

\begin{description}
\item [(A1m)] (Poisson degree distribution) Denote by $N_{k,j}(t)$ the number of class $k$ skills in a text $t$ in the set of class $j$ texts $\calD_j$. Then $\{N_{k,j}(t), t\in \calD_j\}$, $j=1,2, \ldots, J$, $k=1,2, \ldots, K$, are {\em independent} Poisson random variables with means $c_{k,j}$.
\item [(A2m)] (Uniform connection) Each class $(k,j)$-edge of a text node $t\in\calD_j$ is
connected to a skill node in $\calS_k$ independently and uniformly.
Thus, the probability that a class $(k,j)$-edge from a text $t\in\calD_j$ is connected to a particular skill node $s\in\calS_k$ is $1/|\calS_k|$.
\end{description}

From (A1m) and (A2m), the number of edges between a class $j$ text node and a class $k$ skill node follows a Poisson distribution with mean $c_{k,j}/|\yw\calS_k|$. Since there are $\alpha_j |\calD|$ class $j$ text nodes, the number of edges between a class $k$ skill node and class $j$ text nodes is also Poisson distributed with mean $c_{k,j}\alpha_j |\calD|/|\yw\calS_k|=c_{k,j}\frac{\alpha_j}{\beta_k}R$. This leads to the following proposition.

\bprop{skilldegreem}
Consider the sampled skill-text bipartite graph from a semantic language ${\cal L}=({\cal A},{\yw\calT,\calS}, \phi)$ with multiple classes of skills and texts. Then:
\noindent (i) (Poisson degree distribution) Let $M_{k,j}(s)$ be the number of class $j$ text nodes connected to a class $k$ skill node $s$.  Then $\{M_{k,j}(s), s \in \calS_k\}$, $j=1,2, \ldots, J$, $k=1,2, \ldots, K$, are {\em independent} Poisson random variables with means
$d_{k,j}=c_{k,j}\frac{\alpha_j}{\beta_k}R$.
As such, the degree of a class $k$ skill node follows a Poisson distribution with mean
$d_k=\sum_{j=1}^J c_{k,j}\frac{\alpha_j}{\beta_k}R$.

\noindent (ii) (Uniform connection) {\cs  Each  class $(k,j)$-edge of a class $k$ skill node is connected to a class $j$ text node independently and uniformly. Thus,
the probability that a class $(k,j)$-edge from a skill node $s\in\calS_k$ is connected to a particular class $j$ text node is $1/(\alpha_j |\calD|)$.}
\eprop

To gain insight into such an extension, we note that the collection of class $(k,j)$-edges forms a bipartite graph with a single class of skills and a single class of texts. As such, the ensemble of bipartite graphs generated with $K$ classes of skills and $J$ classes of texts can be viewed as the union of $K \times J$ independent ensembles of bipartite graphs, each with a single class of skills and a single class of texts. In \rfig{skilltextmul}, we show an illustration for a skill-text bipartite graph with two classes of skills and three classes of texts.

\begin{figure}[t]
    \centering
    \includegraphics[width=0.9\columnwidth]{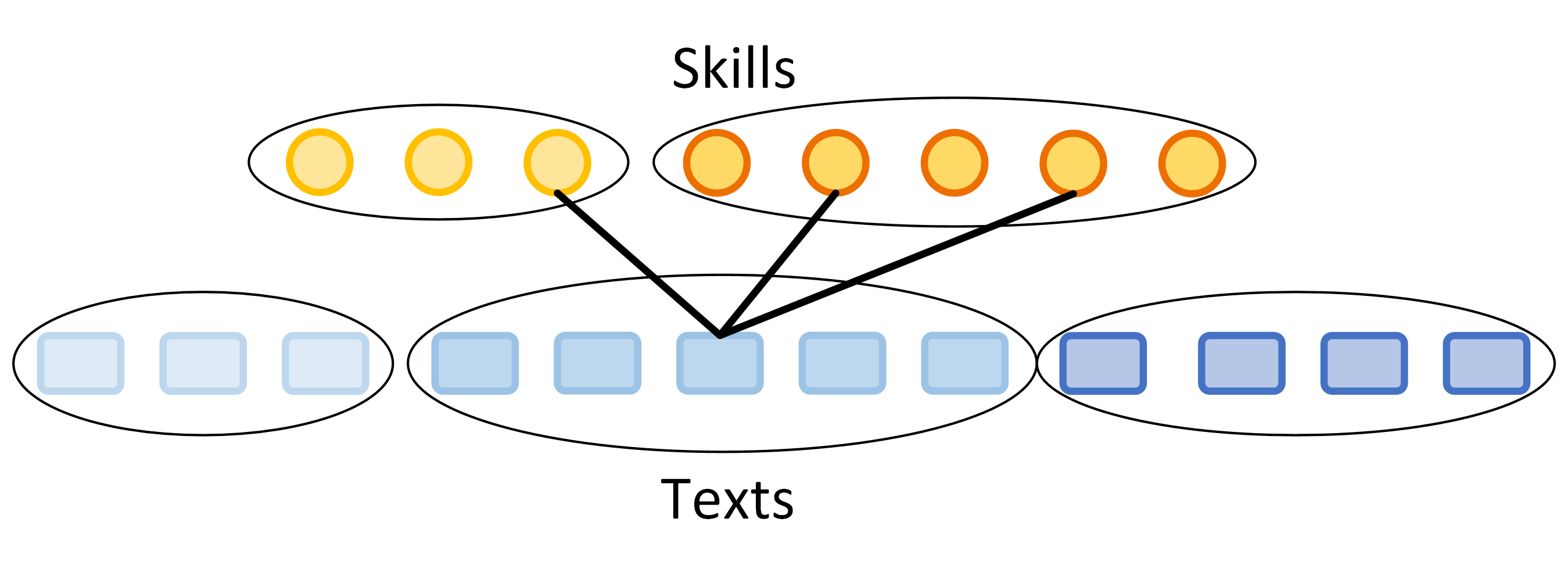}
    \caption{An illustration for a skill-text bipartite graph with two classes of skills and three classes of texts.
    }	
    \label{fig:skilltextmul}
    \end{figure}

An edge is a class $k$ edge if the skill end of the edge is connected to a class $k$ skill.
From \rprop{skilldegreem}, the probability that
a class $k$ edge is a class $(k,j)$-edge is
\bear{skillm2222}
r_{k,j}=\frac{d_{k,j}}{d_k}=\frac{c_{k,j}\alpha_j}{\sum_{j^\prime=1}^{J}c_{k,j^\prime}\alpha_{j^\prime}}.
\eear

\subsubsection{Density evolution for Poisson learners with multiple classes of skills}
\label{sec:Poissonmskill}

 We say a Poisson learner with $K$ classes of skills is subject to a Poisson offered load  $\rho=(\rho_1, \rho_2, \ldots, \rho_K)$ from a text $t$ if  the number of class $k$ skills connected to the text node $t$, $k=1,2, \ldots, K$,  are independent Poisson random variables with means $\rho_k$, $k=1,2, \ldots, K$.
Analogous to the definition of a Poisson receiver with multiple class of input traffic in \cite{chang2020Poisson}, we can define a Poisson learner with multiple class of skills as follows.

\bdefin{Poissonmul}{(\bf Poisson learners with multiple classes of skills)}
	An abstract learner is called a {\em $(P_{{\rm suc},1}(\rhog), P_{{\rm suc},2}(\rhog), \ldots, P_{{\rm suc},K}(\rhog))$-Poisson learner} with $K$ classes of skills if the learner is subject to a Poisson offered load  $\rho=(\rho_1, \rho_2, \ldots, \rho_K)$ from a text, a tagged (randomly selected) class $k$  skill is learned  from the text with probability $P_{{\rm suc},k}(\rhog)$, for $k=1,2, \ldots, K$.
\edefin

 For a $(P_{{\rm suc},1}(\rhog), P_{{\rm suc},2}(\rhog), \ldots, P_{{\rm suc},K}(\rhog))$-Poisson learner,
 the {\em throughput} of class $k$ skills (defined as the expected number of class $k$ skills that are learned) from a text
with a Poisson offered load $\rho$  is thus given by
\beq{Poithrmul}
\Theta_k=\rho_k \cdot P_{{\rm suc},k}(\rhog),
\eeq
$k=1,2, \ldots, K$.

%\subsubsection{Density evolution analysis}
%\label{sec:densityana}

{\cs
Suppose that $J$ classes of texts (a total of $|\calD|$ texts) are presented to a Poisson learner with $K$ classes of skills.
% to learn the skills.
Our study utilizes the tree evaluation approach, as detailed in \cite{luby1998analysis,liva2011graph,paolini2011graph,paolini2012random}, along with the concepts of reduced Poisson offered load, referenced in \cite{chang2020Poisson}. The tree evaluation method is also recognized as the density evolution analysis within the field of information theory \cite{gallager1962low,luby1998analysisb,richardson2001capacity}.}
\iffalse % red
Our study utilizes the tree evaluation approach, as detailed in \cite{luby1998analysis,liva2011graph,paolini2011graph,paolini2012random}, along with the concepts of reduced Poisson offered load, referenced in \cite{chang2020Poisson}. The tree evaluation technique, further developed in \cite{luby1998analysisb,richardson2001capacity}, has been instrumental in determining the capacity of low-density parity-check (LDPC) codes \cite{gallager1962low}. This method is popularly recognized as the density evolution analysis within the field of information theory. {\red [Is the above discussion on density evolution the same as that in previous sections? If so, we can maybe reduce or make more concise.]} Additionally, the reduced Poisson offered load approach is a prominent method for analyzing queueing networks, as exemplified by references such as \cite{kelly2011reversibility,walrand1983probabilistic,kelly1991loss}.
\fi
Similar to the analysis in the original IRSA \cite{liva2011graph} and the coded Poisson receivers \cite{chang2022stability},
%Following \cite{liva2011graph} and \cite{chang2022stability},
we define $\Lambda_{k,\ell}$ as the probability that a class $k$ skill node has $\ell$ edges.
From \rprop{skilldegreem}, we know that
\bear{extm1111}
	\Lambda_{k,\ell}=e^{-d_{k}} \frac{d_k^\ell}{\ell!}, \;\ell=1,2,\dots
\eear
Let
\beq{mean0000mulr}
	\Lambda_{k}(x)=\sum_{\ell=0}^\infty \Lambda_{k,\ell} \cdot x^\ell
\eeq
be the generating function of the {\em degree distribution} of a class $k$ skill node, and
\beq{mean3333mulr}
	\lambda_{k}(x)=\frac{\Lambda_{k}^\prime(x)}{\Lambda_{k}^{\prime}(1)}
\eeq
be the generating function of the {\em excess degree distribution} of a class $k$ skill node.
Since the excess  degree distribution of a Poisson degree distribution is also Poisson with the same mean
(as it can be easily shown by using \req{mean3333mulr}),
we have
\beq{mean3344mulr}
	\Lambda_{k}(x)=\lambda_{k}(x)=e^{-d_{k}(1-x)} .
\eeq

Our density evolution analysis involves the following steps:

\noindent (i)   Let $q_{k}^{(i)}$ be the probability that the {\em skill end} of a randomly selected class $k$ edge has not been learned after the $i^{th}$ SCNS iteration. The offered load of a class $j$ text to the Poisson learner after the $i^{th}$ SCNS iteration has a Poisson distribution with  mean $q_{k}^{(i)} c_{k,j}$.
As pointed out in \cite{chang2020Poisson}, this is due to two important closure properties of Poisson random variables: (a) randomly splitting a Poisson random process yields independent Poisson random processes, and (b) the excess degree distribution of a Poisson random variable is Poisson with the same mean. Consequently, the Poisson offered load is reduced from $c_{k,j}$ to $q_{k}^{(i)} c_{k,j}$ after the $i^{th}$ SCNS iteration.
Let
\beq{rho0000r}
	\tilde c_j=(c_{1,j}, c_{2,j}, \ldots, c_{K,j}),
\eeq
and
\beq{rhoiiiir}
	q^{(i)}=(q_{1}^{(i)}, q_{2}^{(i)}, \ldots, q_{K}^{(i)}).
\eeq
We can represent the offered load at a class $j$ Poisson receiver after the $i^{th}$ SCNS iteration by the vector $q^{(i)} \circ \tilde c_j$, where
$\circ$ denotes the element-wise multiplication of two vectors.
This reduced load argument is also the key step that greatly reduces the computational complexity in the density evolution method.

\noindent (ii) Let $p_{k,j}^{(i+1)}$ be the probability that the {\em text end} of a randomly selected class $(k,j)$-edge has not been learned after the $(i+1)^{th}$ SCNS iteration. Then
\beq{tag6666amulr}
	p_{k,j}^{(i+1)}=1- P_{{\rm suc},k}(q^{(i)} \circ \tilde c_j).
\eeq
%That \req{tag6666amulr} holds
The above equation follows directly from the definition of a Poisson learner in \rdef{Poissonmul} as the offered load of a class $j$ text to the Poisson learner after the $i^{th}$ SCNS iteration is $q^{(i)} \circ \tilde c_j$.

\noindent (iii) Let $p_{k}^{(i+1)}$ be the probability that  the {\em text end} of a randomly selected class $k$ edge has not been learned after the $(i+1)^{th}$ SCNS iteration. Since a class $k$ edge is a class $(k,j)$-edge with probability $r_{k,j}$, it follows that
\bear{tag6666bmulr}
	p_k^{(i+1)}&=&\sum_{j=1}^J r_{k,j}p_{k,j}^{(i+1)} \nonumber\\
	&=&1-\sum_{j=1}^J r_{k,j}P_{{\rm suc},k}(q^{(i)} \circ \tilde c_j).
\eear

\noindent (iv) The probability $q_k^{(i)}$ can be computed recursively from the following equation:
\beq{tag6666cmulr}
	q_{k}^{(i+1)}=\lambda_{k}\Big(1- \sum_{j=1}^J r_{k,j}P_{{\rm suc},k}(q^{(i)} \circ \tilde c_j)\Big),
\eeq
with $q_{k}^{(0)}=1$. To see this, note that a skill end $s$ of a randomly selected edge is not learned after the $(i+1)^{th}$ iteration if the text nodes connected to $s$ (excluding the text end of the randomly selected edge) were not able to learn the skill $s$ after the $i^{th}$ iteration.
Let $\lambda_{k,\ell}$ be the probability that the skill end of a randomly selected class $k$ edge  has additional $\ell$ edges.
 Thus, the probability that the {\em skill} end of a randomly selected class $k$ edge  cannot be learned after the $(i+1)^{th}$ iteration is
\bear{tag2222mulr}
	q_{k}^{(i+1)}&=&1-\sum_{\ell=0}^\infty \lambda_{k,\ell} \cdot \Big (1-(p_{k}^{(i+1)})^{\ell} \Big) \nonumber\\
	&=&\lambda_{k}(p_{k}^{(i+1)}).
\eear
Using \req{tag6666bmulr} in \req{tag2222mulr} yields \req{tag6666cmulr}.
Note from \req{mean3344mulr} and \req{tag2222mulr} that
\beq{skillm3333}
q_{k}^{(i+1)}=e^{-d_{k}(1-p_{k}^{(i+1)})} .
\eeq

\noindent (v) Let $\tilde P_{{\rm suc},k}^{(i)}$ be the probability that a  randomly selected {\em class $k$ skill} can be learned after the $i^{th}$ iteration. This is the probability that at least one text end of the edges connected to $s$ has been learned after the $i^{th}$ iteration. Since the probability that a randomly selected {\em class $k$ skill} node has  $\ell$ edges is $\Lambda_{k,\ell}$, we have from \req{tag6666bmulr} that
\begin{align}
&\tilde P_{{\rm suc},k}^{(i)}=\sum_{\ell=0}^\infty \Lambda_{k,\ell} \cdot \Big (1-(p_{k}^{(i)})^{\ell} \Big)\nonumber\\
	&=\sum_{\ell=0}^\infty \Lambda_{k,\ell} \cdot \Big (1-(1- \sum_{j=1}^J r_{k,j}P_{{\rm suc},k}(q^{(i-1)} \circ \tilde c_j ))^{\ell} \Big) \nonumber \\
	&=1-\Lambda_k\Big (1- \sum_{j=1}^J r_{k,j}P_{{\rm suc},k}(q^{(i-1)} \circ \tilde c_j)\Big).\label{eq:mean5555dmulr}
\end{align}

Here, we only outline the steps to derive the recursive equations. A rigorous proof for the density evolution analysis requires the concentration theorem (Theorem 2 of \cite{richardson2001capacity}) to prove the ergodicity of the system.
i.e., as $|\calD| \to \infty$, the average fraction of skill nodes that have not been learned after the $i^{th}$ SCNS iteration converges to the probability that a randomly selected skill node has not been learned after the $i^{th}$ SCNS iteration.

One can also represent the recursive equations using $p_k^{(i)}$'s.
From \req{tag6666bmulr} and \req{skillm3333}, we also have for $k=1,2, \ldots, K$,
\bear{skillm4444}
p_{k}^{(i+1)}=1-\sum_{j=1}^J r_{k,j}P_{{\rm suc},k}(\gamma_1^{(i)}, \gamma_2^{(i)}, \ldots, \gamma_K^{(i)}).
\eear
where
\beq{skillm5555}
\gamma_k^{(i)}=e^{-d_{k}(1-p_{k}^{(i)})}c_{k,j},
\eeq
$k=1,2, \ldots, K$.
These $K$ coupled equations can be used for obtaining the scaling law for $p$ with respect to $R$, given the parameters $c_{k,j}$'s, $\alpha_j$'s, and $\beta_k$'s. Note that \req{skillm4444} reduces to \req{one6666} when $J=K=1$ and $i \to \infty$.

\bsubsec{Deterministic $\psi$-learners}{dlearners}

The framework of Poisson learners operates within a probabilistic context where skills are learned according to certain probabilities.
As in \cite{liu2021aloha}, we can also define deterministic learners.
 Denote by ${\cal Z}^+$  the set of nonnegative integers. We say a learner with $K$ classes of skills is subject to a {\em deterministic} load   $n=(n_1, n_2, \ldots, n_K)  \in {{\cal Z}^+}^K$ from a text if the number of class $k$ skills is $n_k$.

\bdefin{phiALOHA}{\bf ($\psi$-learner with multiple classes of skills \cite{liu2021aloha})}
    Consider a  deterministic function $$\psi: {{\cal Z}^+}^K \rightarrow {{\cal Z}^+}^K$$ that maps a $K$-vector $n=(n_1, n_2, \ldots, n_K)$ to the $K$-vector $(\psi_1(n), \psi_2(n), \ldots, \psi_K(n))$. An abstract learner is called a $\psi$-learner (with $K$ classes of skills) if  the number of class $k$ skills that are learned is exactly $\psi_k(n)$, $k=1,2, \ldots, K$, when the learner is subject to a deterministic load $n=(n_1, n_2, \ldots, n_K)$.
    \edefin

One typical example of the $\psi$-learner is 1-skill learner. As before, one can also generalize  1-skill learner to
$|\calD|$-skill learner, where there are at most $|\calD|$ novel skills can be learned.
The $|\calD|$-skill learner is a $\psi$-learner with
\beq{phi7722}
    \psi(n)=\left \{\begin{array}{cc}
        n & \mbox{if}\; n \le D \\
        0 & \mbox{otherwise}
    \end{array} \right ..
\eeq

Analogous to the proof of Theorem 14 in \cite{liu2021aloha}, it is shown that for every $\psi$-learner, there is an induced Poisson learner. This relationship is established by computing the throughput of the $\psi$-learner when it is subject to a Poisson offered load $\rho$, and then using \req{Poithrmul} to determine the success probability function. For the $|\calD|$-skill learner, the throughput for a Poisson offered load $\rho$ is given by
$$ \sum_{t=1}^{D} t\frac{e^{-\rho} \rho^{t}}{t!}=\rho \sum_{t=0}^{D-1} \frac{e^{-\rho} \rho^{t}}{t!}.$$
From \req{Poithrmul}, it is a Poisson learner with the following success probability function:
\beq{tfold1111}
    P_{\rm suc}(\rhog)=\sum_{t=0}^{D-1} \frac{e^{-\rho} \rho^{t}}{t!}.
\eeq

To motivate the extension to multiple classes of skills, consider a learner with two classes of skills: class 1 and class 2. Similar to the 1-skill learner, this learner can learn a novel skill, whether class 1 or class 2, if it is the only novel skill in a text. Additionally, when a text contains both a class 1 novel skill and a class 2 novel skill, the learner can first learn the class 1 skill and then apply the SCNS strategy to reduce the number of novel skills to a single class 2 skill. Consequently, since the learner is capable of learning a single class 2 novel skill, it can also learn the class 2 skill in the text.
Let $n_1$ and $n_2$
be the number of class 1 and class 2 skills, respectively, in a text. Then, such a deterministic learner is a $\psi$-learner with two classes of skills, where
\beq{phi7724}
   \psi(n_1,n_2)=\left \{\begin{array}{cc}
       (n_1,n_2) & \mbox{if}\; (n_1,n_2) \le (1,1) \\
       (0,0) & \mbox{otherwise}
   \end{array} \right ..
\eeq

Since such a learning process is analogous to the near-far successive interference cancellation decoding scheme in \cite{ordentlich2017low},
we refer to such a $\psi$-learner as a near-far learner.
Now we can use the throughput formula in \req{Poithrmul} to show that a near-far learner
is also a Poisson learner with  the following two success probability functions:
\bear{phi7729}
P_{{\rm suc},1}(\rho_1,\rho_2)&=&e^{-\rho_1}(e^{-\rho_2}+\rho_2 e^{-\rho_2}), \nonumber\\
P_{{\rm suc},2}(\rho_1,\rho_2)&=&e^{-\rho_2}(e^{-\rho_1}+\rho_1 e^{-\rho_1}).
\eear

Suppose that there is only one class of texts, i.e., $J=1$. Now consider training with the near-far learner.
Then we have from \req{skillm4444}
that
\bear{skillm9900}
&&p_1^{(i+1)}=1- P_{{\rm suc},1}(e^{-d_1 (1-p_1^{(i)})}c_1, e^{-d_2 (1-p_2^{(i)})}c_2) ,\nonumber\\
&&p_2^{(i+1)}=1- P_{{\rm suc},2}(e^{-d_1 (1-p_1^{(i)})}c_1, e^{-d_2 (1-p_2^{(i)})}c_2), \nonumber\\
\eear
where $d_k=(c_1+c_2)R/\beta_k$, $k=1$ and 2.

\begin{figure}[t]
	\centering
	\includegraphics[width=0.48\textwidth]{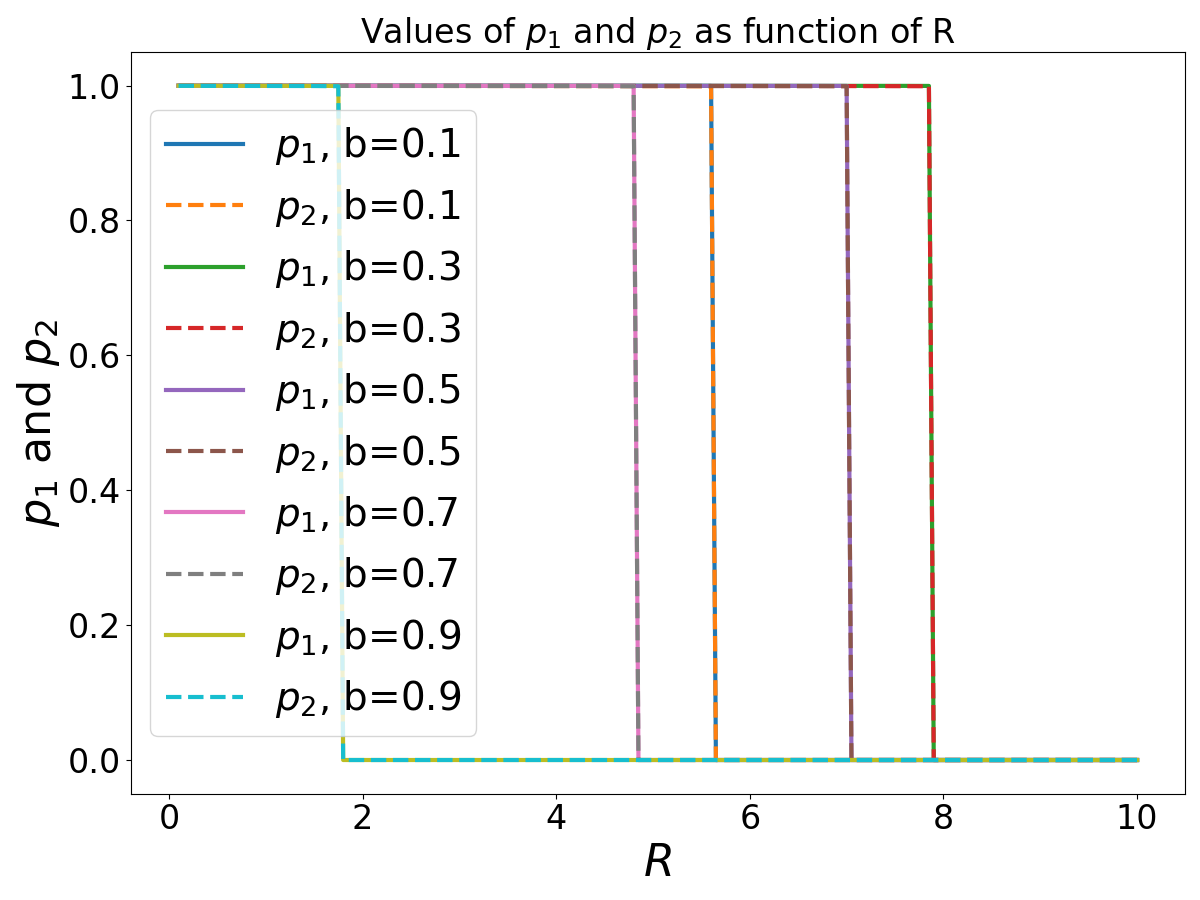}
	\caption{The plot illustrates the computed values of \( p_1 \) (and \( p_2 \)) for the near-far learner as a function of \( R \), which ranges from 0.1 to 10 with a step size of 0.05, for each \( \beta \) value (0.1, 0.3, 0.5, 0.7, 0.9). The average degrees of the two classes of skills are $c_1=3$ and $c_2=7$, respectively. The computation is based on the two recursive equations in \req{skillm9900} with the initial condition \( p_1^{(0)} =p_2^{(0)}= 1 \), where $P_{{\rm suc},k}(\rho)$, $k=1$ and 2, are defined in \req{phi7729}. The number of iterations is 100.}
	\label{fig:ptwoclass}
\end{figure}

In \rfig{ptwoclass}, we plot the computed values of \( p_1 \) as a function of \( R \) for each \( \beta \) value (0.1, 0.3, 0.5, 0.7, 0.9). The average degrees of the two classes of skills are \(c_1=3\) and \(c_2=7\), respectively. For this example, it is noted that the values of \( p_2 \) overlap with those of \( p_1 \), making \( p_2 \) indistinguishable in the figure. Interestingly, thresholds are also observed for these five curves. As demonstrated in \cite{chang2022stability}, the thresholds for these two classes of skills coincide due to the coupling of the density evolution equations (similar to those in \req{skillm9900}), leading to the simultaneous emergence of various classes of skills.

\bsec{Application to Semantic Compression and Communication}{compression}

The lossless compression of a text requires exact recovery of the sequence of tokens in a text. However, if our interest lies in recovering only the semantic meaning of a text, we might be able to compress it using fewer bits than required for lossless compression. This concept is referred to as the semantic (level B) problem in \cite{weaver1953recent}.

\bdefin{compression}({\bf Semantic Compression}) A compression method is termed a semantic compression method if the recovered text is semantically equivalent to the original text. An abstract learner is called {\em generative} if it can generate a text of tokens given a set of learned skills.
\edefin

In this section, we assume that the abstract learners discussed in the previous section are also generative. Intuitively, to render an abstract learner generative, one can store a ``table'' of tokens corresponding to learned skills. Given a set of learned skills, the generative learner then uses the stored tokens to compose a text. It is known from \cite{ramsauer2020hopfield} that transformers \cite{vaswani2017attention} are, in fact, Hopfield networks, which are a form of associative memory. In other words, transformers are capable of storing tables of tokens for learned skills.

Here, we demonstrate how to use {\em generative} learners for semantic compression. Once the training of an abstract learner
is complete, the expected number of skills learned is $|{\yw \calS}|(1-e^{-c R(1-p)})$
with $p$ given by \req{one3333} for the 1-skill learner and by \req{one6666} for the Poisson learner. By indexing the learned skills,
the number of bits required to represent a learned skill is $\log_2 (|{\yw \calS}|(1-e^{-c R(1-p)}))$. Therefore, if a text with $k$ skills is understood by the learner, it can be compressed/encoded using $k \log_2 (|{\yw \calS}|(1-e^{-c R(1-p)}))$ bits. As previously shown, the probability that a randomly selected text is {\em understood} by the learner is $e^{-( c e^{-c R(1-p)} )}$, and it requires on average $c\log_2 (|{\yw \calS}|(1-e^{-c R(1-p)}))$ bits to encode the text. Conversely, if a randomly selected text is {\em not understood} by the learner, it can be encoded using a lossless compression encoder. Suppose the lossless compression encoder requires, on average, $z$ bits to compress a text. Then the semantic compression method described above requires, on average,
$e^{-( c e^{-c R(1-p)} )}c\log_2 (|{\yw \calS}|(1-e^{-c R(1-p)}))
+ (1-e^{-( c e^{-c R(1-p)} )})z$
bits for a text. By Shannon's analysis in \cite{shannon1951english}, the entropy per word in the English language is approximately $11.82$ bits. By assuming that an average sentence has approximately $20$ words, the number of bits required to represent a text is approximately $z=236.4$ bits. With $R$ sufficiently large such that $c\log_2 (|{\yw \calS}|(1-e^{-c R(1-p)}))\approx c\log_2 |{\yw \calS}|$ and with $c$ set as $5$, a compression gain is obtained as long as $|{\yw \calS}|\leq 2^{47.28}$.

For semantic communication, we can utilize a semantic encoder/decoder. In \rfig{block}, we present the general architecture of a semantic communication system, which consists of a semantic transmitter, a semantic receiver, and a physical channel. The semantic transmitter includes a semantic encoder and a channel encoder, while the semantic receiver includes a semantic decoder and a channel decoder. The communication between the channel encoder and decoder occurs over the physical channel, referred to as level A communication in \cite{weaver1953recent}. {\cs Conversely, the communication between the semantic encoder and decoder
is termed level B communication in \cite{weaver1953recent}.}

Recent literature on semantic communications, such as \cite{deepsc2021,UTdeepsc2022,deepscvqvae2023}, has proposed an end-to-end approach for jointly training the semantic and channel encoder/decoder. This method is claimed to be superior to separate training. However, the end-to-end approach does not scale efficiently with data size. {\yw In fact, as shown in the previous section, a large dataset is necessary for training to exhibit the emergence of semantic capability. Hence, for the transmission of texts from a general semantic language, an LLM model is required at both the semantic transmitter and receiver, which would be difficult to retrain and adapt to varying physical channels. In light of this, a modular design (i.e., separate source and channel coding) may be more effective for semantic communication in practice.}

\begin{figure}[t]
    \centering
    \includegraphics[width=0.8\columnwidth]{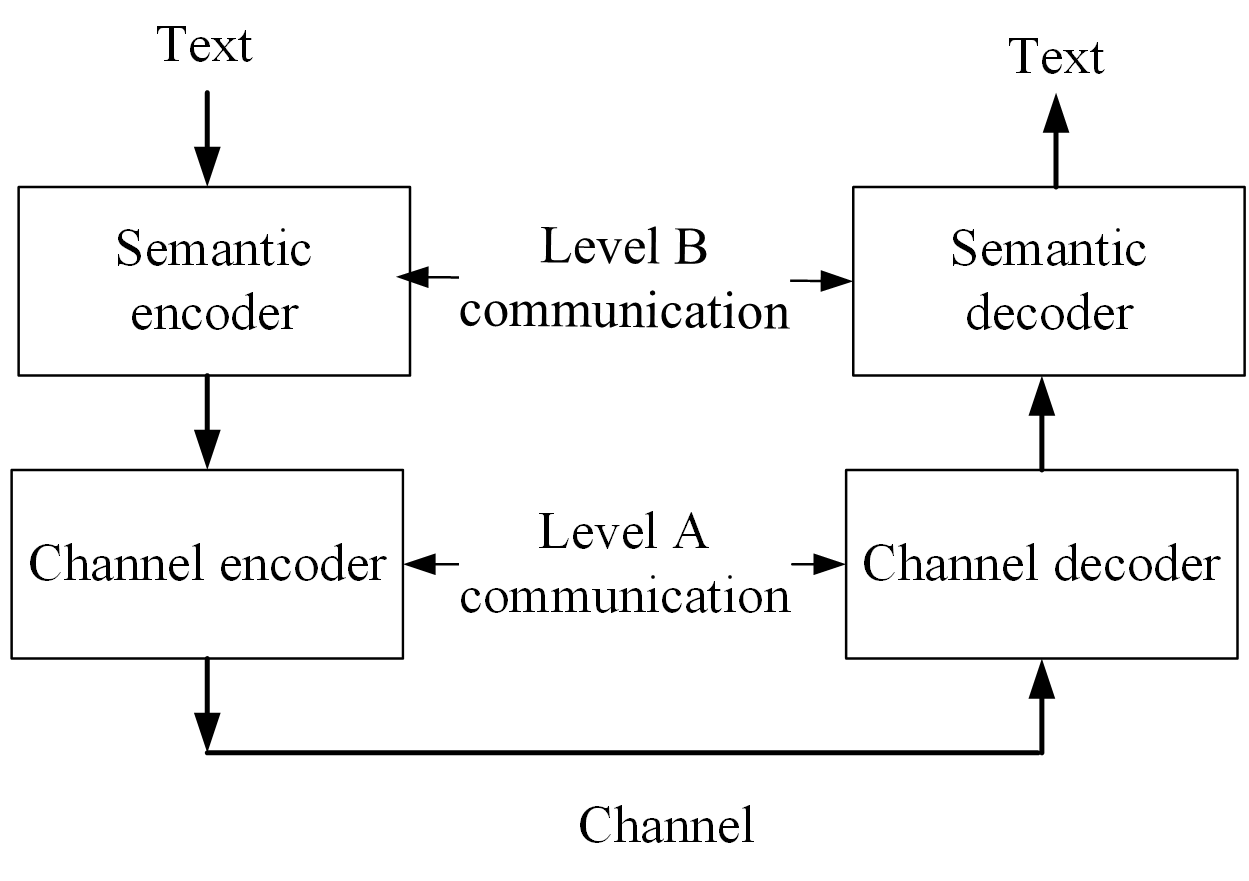}
    \caption{The general framework of a semantic communication system.}	
    \label{fig:block}
      \end{figure}

For domain-specific texts discussed in \rsec{hskills}, semantic compression can also be used
once fine-tuning is complete. If the number of domain-specific skills is much smaller than the number of basic skills, i.e., $|\yw \calS_f| \ll |\calS|$, one expects the compression gain for domain-specific texts in the fine-tuning model will surpass that achieved using the foundation model.

\bsec{Conclusions and Future Directions}{con}
%%%%%%%%%%%%%%%%%%%%%%%%%%%%%%%%%%%%%%%%%%%%%%%%%%%%%%%%%%%%%%%%%%

Inspired by \cite{arora2023theory}, we defined a semantic language as a skill-text bipartite graph. Skills can be learned by repeatedly presenting training texts to abstract learners, such as 1-skill learners and Poisson learners. By using density evolution analysis, we showed the emergence of learned skills when the ratio $R$ exceeds a certain threshold. Moreover, we derived the corresponding scaling law for the testing error. We also demonstrated how trained abstract learners can be used for semantic compression and semantic communication.

The present work is intended to inspire further thoughts on the emergence of learned skills in today's LLMs, drawing relation to the iterative decoding procedure in communications and information theory. However, we recognize that
there are still significant gaps between our theory and real-world systems.
Several key factors that should be further investigated include:
%to better link theory with practice are listed below:

\begin{itemize}
\item[(i)] Modeling abstract learners: Real-world LLMs, such as GPT-4 \cite{GPT4} and Gemini \cite{pichai2023introducing}, are transformer-based and are trained using the gradient descent algorithm. It is not clear how a {\em skill} is learned when a text of tokens is fed into these transformers.

%\noindent (ii) Connections to Kolmogorov complexity:
%In a recent talk \cite{Sutskever2023}, Sutskever pointed out the connection between compression and unsupervised learning through Kolmogorov complexity. It appears that GPT-4 has learned some structure of languages and uses that structure for semantic compression. However, it is not clear how such a structure is represented and learned.

\iffalse
\item[(ii)] Modeling the structure of a semantic language: In this paper, we use a skill-text bipartite graph to model the structure of a semantic language. While this may be sufficient for a machine's interpretation of semantics for compression, this model is overly simplistic for human's interpretation of real-world languages, which often have a hierarchical structure of skills.
    %For example, basic skills must be learned before a complex skill can be mastered. Even though density evolution might be extended to settings with hierarchical skill structures, deriving simple scaling laws like those presented in this paper is  significantly more challenging.
   % unlikely.
\fi

\item[(ii)] Order of skills: If we view a skill as a {\em thought}, often a text can be represented by a chain (sequence) of thoughts.
As such, the order of the skills appearing in a text could also be an important factor,
thus, learning the mapping $\phi$ alone might not suffice for effective semantic encoding/decoding.

\item[(iii)] Soft learning: In our model for an abstract learner, a skill is learned in a definite manner. Exploring a soft learning setting, where a skill is {\em partially learned}, could be beneficial. For transformer-based LLMs, a softmax function is used for the next token prediction. Extending our theory to encompass soft learning might help bridge the gap between theory and real-world systems.

\item[(iv)] Scaling laws of the number of parameters: The scaling law obtained in this paper only pertains to the size of training texts. It is unclear how the testing error scale with the number of parameters. For instance, GPT-3 is known to have more than 1,750 billion parameters, and LLMs prior to GPT-3 do not exhibit the emergence of skills.
One possible explanation for the necessity of a larger number of parameters is that the capacity of a transformer, defined as the number of skills (or patterns) that can be stored in the corresponding Hopfield network \cite{ramsauer2020hopfield}, depends on the number of parameters in the transformer.

\item[(v)] Experimental verification: Performing experiments to verify our theory in the context of real-world semantic languages is difficult. However, it may
be possible to construct a semantic language, e.g., a simple programming language, with a small number of skills and employ transformer-based LLMs with a relatively small number of parameters to verify the theory.

\end{itemize}

%\newpage

%%%%%%%%%%%%%%%%%%%%%%%%%%%%%%%%%%%%%%%%%%%%%%%%%%
%                Bibliography                    %
%%%%%%%%%%%%%%%%%%%%%%%%%%%%%%%%%%%%%%%%%%%%%%%%%%
%\bibliographystyle{IEEEtran}
%\bibliography{bibdatabase,mobshort,Phase}

\begin{thebibliography}{10}
\providecommand{\url}[1]{#1}
\csname url@samestyle\endcsname
\providecommand{\newblock}{\relax}
\providecommand{\bibinfo}[2]{#2}
\providecommand{\BIBentrySTDinterwordspacing}{\spaceskip=0pt\relax}
\providecommand{\BIBentryALTinterwordstretchfactor}{4}
\providecommand{\BIBentryALTinterwordspacing}{\spaceskip=\fontdimen2\font plus
\BIBentryALTinterwordstretchfactor\fontdimen3\font minus
  \fontdimen4\font\relax}
\providecommand{\BIBforeignlanguage}[2]{{%
\expandafter\ifx\csname l@#1\endcsname\relax
\typeout{** WARNING: IEEEtran.bst: No hyphenation pattern has been}%
\typeout{** loaded for the language `#1'. Using the pattern for}%
\typeout{** the default language instead.}%
\else
\language=\csname l@#1\endcsname
\fi
#2}}
\providecommand{\BIBdecl}{\relax}
\BIBdecl

\bibitem{GPT4}
OpenAI, ``{GPT}-4 technical report,''
  \emph{https://cdn.openai.com/papers/gpt-4.pdf}, 2023.

\bibitem{pichai2023introducing}
S.~Pichai and D.~Hassabis, ``Introducing gemini: our largest and most capable
  {AI} model,'' \emph{Google. Retrieved December}, 2023.

\bibitem{devlin2018bert}
J.~Devlin, M.-W. Chang, K.~Lee, and K.~Toutanova, ``{BERT}: Pre-training of
  deep bidirectional transformers for language understanding,'' \emph{arXiv
  preprint arXiv:1810.04805}, 2018.

\bibitem{brown2020language}
T.~Brown, B.~Mann, N.~Ryder, M.~Subbiah, J.~D. Kaplan, P.~Dhariwal,
  A.~Neelakantan, P.~Shyam, G.~Sastry, A.~Askell \emph{et~al.}, ``Language
  models are few-shot learners,'' \emph{Advances in Neural Information
  Processing Systems}, vol.~33, pp. 1877--1901, 2020.

\bibitem{kaplan2020scaling}
J.~Kaplan, S.~McCandlish, T.~Henighan, T.~B. Brown, B.~Chess, R.~Child,
  S.~Gray, A.~Radford, J.~Wu, and D.~Amodei, ``Scaling laws for neural language
  models,'' \emph{arXiv preprint arXiv:2001.08361}, 2020.

\bibitem{hoffmann2022training}
J.~Hoffmann, S.~Borgeaud, A.~Mensch, E.~Buchatskaya, T.~Cai, E.~Rutherford,
  D.~d.~L. Casas, L.~A. Hendricks, J.~Welbl, A.~Clark \emph{et~al.}, ``Training
  compute-optimal large language models,'' \emph{arXiv preprint
  arXiv:2203.15556}, 2022.

\bibitem{wei2022emergent}
J.~Wei, Y.~Tay, R.~Bommasani, C.~Raffel, B.~Zoph, S.~Borgeaud, D.~Yogatama,
  M.~Bosma, D.~Zhou, D.~Metzler \emph{et~al.}, ``Emergent abilities of large
  language models,'' \emph{arXiv preprint arXiv:2206.07682}, 2022.

\bibitem{wei2022inverse}
J.~Wei, Y.~Tay, and Q.~V. Le, ``Inverse scaling can become u-shaped,''
  \emph{arXiv preprint arXiv:2211.02011}, 2022.

\bibitem{Newman2010}
M.~Newman, \emph{Networks: an introduction}.\hskip 1em plus 0.5em minus
  0.4em\relax OUP Oxford, 2009.

\bibitem{chang2023simple}
C.-S. Chang, ``A simple explanation for the phase transition in large language
  models with list decoding,'' \emph{arXiv preprint arXiv:2303.13112}, 2023.

\bibitem{vaswani2017attention}
A.~Vaswani, N.~Shazeer, N.~Parmar, J.~Uszkoreit, L.~Jones, A.~N. Gomez,
  {\L}.~Kaiser, and I.~Polosukhin, ``Attention is all you need,''
  \emph{Advances in Neural Information Processing Systems}, vol.~30, 2017.

\bibitem{ramsauer2020hopfield}
H.~Ramsauer, B.~Sch{\"a}fl, J.~Lehner, P.~Seidl, M.~Widrich, T.~Adler,
  L.~Gruber, M.~Holzleitner, M.~Pavlovi{\'c}, G.~K. Sandve \emph{et~al.},
  ``Hopfield networks is all you need,'' \emph{arXiv preprint
  arXiv:2008.02217}, 2020.

\bibitem{arora2023theory}
S.~Arora and A.~Goyal, ``A theory for emergence of complex skills in language
  models,'' \emph{arXiv preprint arXiv:2307.15936}, 2023.

\bibitem{gallager1962low}
R.~Gallager, ``Low-density parity-check codes,'' \emph{IRE Transactions on
  Information Theory}, vol.~8, no.~1, pp. 21--28, 1962.

\bibitem{shokrollahi1999new}
M.~A. Shokrollahi, ``New sequences of linear time erasure codes approaching the
  channel capacity,'' in \emph{International Symposium on Applied Algebra,
  Algebraic Algorithms, and Error-Correcting Codes}.\hskip 1em plus 0.5em minus
  0.4em\relax Springer, 1999, pp. 65--76.

\bibitem{richardson2001design}
T.~J. Richardson, M.~A. Shokrollahi, and R.~L. Urbanke, ``Design of
  capacity-approaching irregular low-density parity-check codes,'' \emph{IEEE
  Transactions on Information Theory}, vol.~47, no.~2, pp. 619--637, 2001.

\bibitem{liva2011graph}
G.~Liva, ``Graph-based analysis and optimization of contention resolution
  diversity slotted {ALOHA},'' \emph{IEEE Transactions on Communications},
  vol.~59, no.~2, pp. 477--487, 2011.

\bibitem{narayanan2012iterative}
K.~R. Narayanan and H.~D. Pfister, ``Iterative collision resolution for slotted
  {ALOHA}: An optimal uncoordinated transmission policy,'' in \emph{Turbo Codes
  and Iterative Information Processing (ISTC), 2012 7th International Symposium
  on}.\hskip 1em plus 0.5em minus 0.4em\relax IEEE, 2012, pp. 136--139.

\bibitem{paolini2012random}
E.~Paolini, G.~Liva, and M.~Chiani, ``Random access on graphs: A survey and new
  results,'' in \emph{Signals, Systems and Computers (ASILOMAR), 2012
  Conference Record of the Forty Sixth Asilomar Conference on}.\hskip 1em plus
  0.5em minus 0.4em\relax IEEE, 2012, pp. 1743--1747.

\bibitem{jakovetic2015cooperative}
D.~Jakoveti{\'c}, D.~Bajovi{\'c}, D.~Vukobratovi{\'c}, and V.~Crnojevi{\'c},
  ``Cooperative slotted {ALOHA} for multi-base station systems,'' \emph{IEEE
  Transactions on Communications}, vol.~63, no.~4, pp. 1443--1456, 2015.

\bibitem{stefanovic2018coded}
{\v{C}}.~Stefanovi{\'c} and D.~Vukobratovi{\'c}, ``Coded random access,'' in
  \emph{Network Coding and Subspace Designs}.\hskip 1em plus 0.5em minus
  0.4em\relax Springer, 2018, pp. 339--359.

\bibitem{chiang2022parallel}
Y.-H. Chiang, Y.-J. Lin, C.-S. Chang, and Y.-W.~P. Hong, ``Parallel decoding of
  irsa with noise,'' in \emph{2022 IEEE 33rd Annual International Symposium on
  Personal, Indoor and Mobile Radio Communications (PIMRC)}.\hskip 1em plus
  0.5em minus 0.4em\relax IEEE, 2022, pp. 320--326.

\bibitem{luby1998analysis}
M.~Luby, M.~Mitzenmacher, and M.~A. Shokrollahi, ``Analysis of random processes
  via and-or tree evaluation,'' in \emph{SODA}, vol.~98, 1998, pp. 364--373.

\bibitem{luby1998analysisb}
M.~Luby, M.~Mitzenmacher, A.~Shokrollah, and D.~Spielman, ``Analysis of low
  density codes and improved designs using irregular graphs,'' in
  \emph{Proceedings of the thirtieth annual ACM symposium on Theory of
  computing}, 1998, pp. 249--258.

\bibitem{richardson2001capacity}
T.~J. Richardson and R.~L. Urbanke, ``The capacity of low-density parity-check
  codes under message-passing decoding,'' \emph{IEEE Transactions on
  Information Theory}, vol.~47, no.~2, pp. 599--618, 2001.

\bibitem{chang2022stability}
C.-M. Chang, Y.-J. Lin, C.-S. Chang, and D.-S. Lee, ``On the stability regions
  of coded {Poisson} receivers with multiple classes of users and receivers,''
  \emph{IEEE/ACM Transactions on Networking}, vol.~31, no.~1, pp. 234--247,
  2022.

\bibitem{chang2020Poisson}
C.-H. Yu, L.~Huang, C.-S. Chang, and D.-S. Lee, ``Poisson receivers: a
  probabilistic framework for analyzing coded random access,'' \emph{IEEE/ACM
  Transactions on Networking}, vol.~29, no.~2, pp. 862--875, 2021.

\bibitem{liu2021aloha}
T.-H. Liu, C.-H. Yu, Y.-J. Lin, C.-M. Chang, C.-S. Chang, and D.-S. Lee,
  ``{ALOHA} receivers: a network calculus approach for analyzing coded multiple
  access with {SIC},'' \emph{IEEE/ACM Transactions on Networking}, vol.~29,
  no.~2, pp. 862--875, 2021.

\bibitem{weaver1953recent}
W.~Weaver, ``Recent contributions to the mathematical theory of
  communication,'' \emph{ETC: a review of general semantics}, pp. 261--281,
  1953.

\bibitem{shannon1951english}
C.~E. Shannon, ``Prediction and entropy of printed english,'' \emph{The Bell
  System Technical Journal}, vol.~30, no.~1, pp. 50--64, 1951.

\bibitem{deepsc2021}
H.~Xie, Z.~Qin, G.~Y. Li, and B.-H. Juang, ``Deep learning enabled semantic
  communication systems,'' \emph{IEEE Transactions on Signal Processing},
  vol.~69, pp. 2663--2675, 2021.

\bibitem{UTdeepsc2022}
Q.~Zhou, R.~Li, Z.~Zhao, C.~Peng, and H.~Zhang, ``Semantic communication with
  adaptive universal transformer,'' \emph{IEEE Wireless Communications
  Letters}, vol.~11, no.~3, pp. 453--457, 2022.

\bibitem{deepscvqvae2023}
Q.~Hu, G.~Zhang, Z.~Qin, Y.~Cai, G.~Yu, and G.~Y. Li, ``Robust semantic
  communications with masked {VQ-VAE} enabled codebook,'' \emph{IEEE
  Transactions on Wireless Communications}, pp. 1--1, 2023.

\bibitem{paolini2011graph}
E.~Paolini, G.~Liva, and M.~Chiani, ``Graph-based random access for the
  collision channel without feedback: Capacity bound,'' in \emph{2011 IEEE
  Global Telecommunications Conference-GLOBECOM 2011}.\hskip 1em plus 0.5em
  minus 0.4em\relax IEEE, 2011, pp. 1--5.

\bibitem{kelly2011reversibility}
F.~P. Kelly, \emph{Reversibility and stochastic networks}.\hskip 1em plus 0.5em
  minus 0.4em\relax Cambridge University Press, 2011.

\bibitem{walrand1983probabilistic}
J.~Walrand, ``A probabilistic look at networks of quasi-reversible queues,''
  \emph{IEEE Transactions on Information Theory}, no.~6, pp. 825--831, 1983.

\bibitem{kelly1991loss}
F.~P. Kelly, ``Loss networks,'' \emph{The annals of applied probability}, pp.
  319--378, 1991.

\bibitem{ordentlich2017low}
O.~Ordentlich and Y.~Polyanskiy, ``Low complexity schemes for the random access
  {Gaussian} channel,'' in \emph{2017 IEEE International Symposium on
  Information Theory (ISIT)}.\hskip 1em plus 0.5em minus 0.4em\relax IEEE,
  2017, pp. 2528--2532.

\end{thebibliography}

%\addtolength{\rightmargin}{0.05in}
%\addtolength{\textwidth}{-0.01in}

\iffalse

\fi


\begin{IEEEbiography}[{\includegraphics[width=1in,height=1.25in,clip,keepaspectratio]{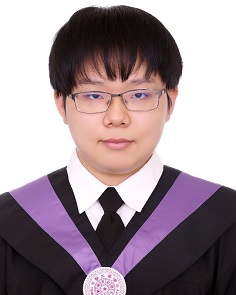}}]
	{Kuo-Yu Liao}
received the B.S. degree in mathematics in 2022 from National Tsing Hua University. He is currently pursuing the M.S. degree in the Institute of Communications Engineering, National Tsing Hua University, Hsinchu, Taiwan.
His research interest is in 5G and beyond wireless communication.
\end{IEEEbiography}

\begin{IEEEbiography}[{\includegraphics[width=1in,height=1.25in,clip,keepaspectratio]{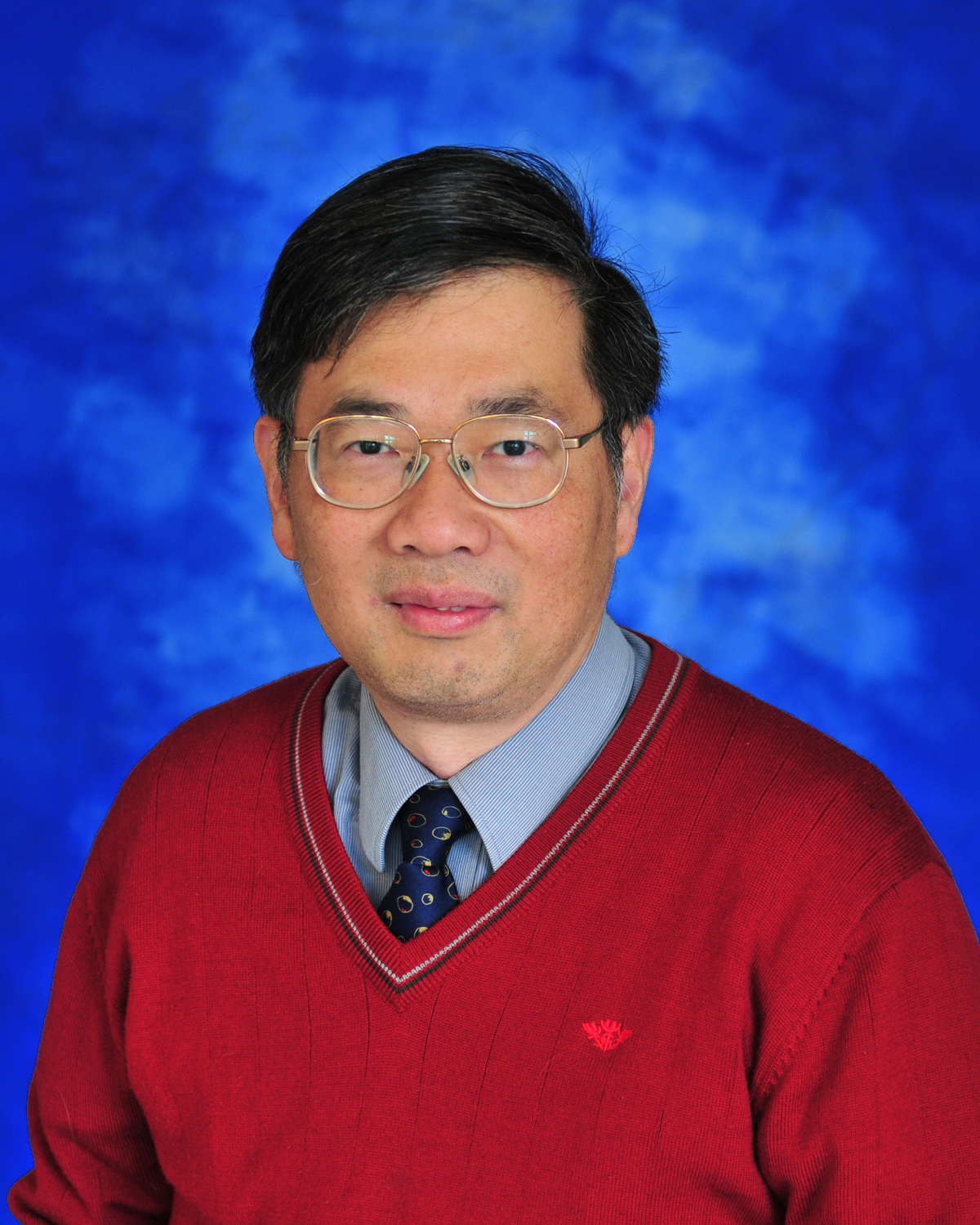}}]
	{Cheng-Shang Chang}
	(S'85-M'86-M'89-SM'93-F'04)
	received the B.S. degree from National Taiwan
	University, Taipei, Taiwan, in 1983, and the M.S.
	and Ph.D. degrees from Columbia University, New
	York, NY, USA, in 1986 and 1989, respectively, all
	in electrical engineering.
	
	From 1989 to 1993, he was employed as a
	Research Staff Member with the IBM Thomas J.
	Watson Research Center, Yorktown Heights, NY,
	USA. Since 1993, he has been with the Department
	of Electrical Engineering, National Tsing Hua
	University, Taiwan, where he is a Tsing Hua Distinguished Chair Professor. He is the author
	of the book Performance Guarantees in Communication Networks (Springer,
	2000) and the coauthor of the book Principles, Architectures and Mathematical
	Theory of High Performance Packet Switches (Ministry of Education, R.O.C.,
	2006). His current research interests are concerned with network science, big data analytics,
	mathematical modeling of the Internet, and high-speed switching.
	
	Dr. Chang served as an Editor for Operations Research from 1992 to 1999,
	an Editor for the {\em IEEE/ACM TRANSACTIONS ON NETWORKING} from 2007
	to 2009, and an Editor for the {\em IEEE TRANSACTIONS
		ON NETWORK SCIENCE AND ENGINEERING} from 2014 to 2017. He is currently serving as an Editor-at-Large for the {\em IEEE/ACM
		TRANSACTIONS ON NETWORKING}. He is a member of IFIP Working
	Group 7.3. He received an IBM Outstanding Innovation Award in 1992, an
	IBM Faculty Partnership Award in 2001, and Outstanding Research Awards
	from the National Science Council, Taiwan, in 1998, 2000, and 2002, respectively.
	He also received Outstanding Teaching Awards from both the College
	of EECS and the university itself in 2003. He was appointed as the first Y. Z.
	Hsu Scientific Chair Professor in 2002. He received the Merit NSC Research Fellow Award from the
	National Science Council, R.O.C. in 2011. He also received the Academic Award in 2011 and the National Chair Professorship in 2017 and 2023 from
	the Ministry of Education, R.O.C. He is the recipient of the 2017 IEEE INFOCOM Achievement Award.
\end{IEEEbiography}

\begin{IEEEbiography}
	[{\includegraphics[width=1in,height=1.25in,clip,keepaspectratio]{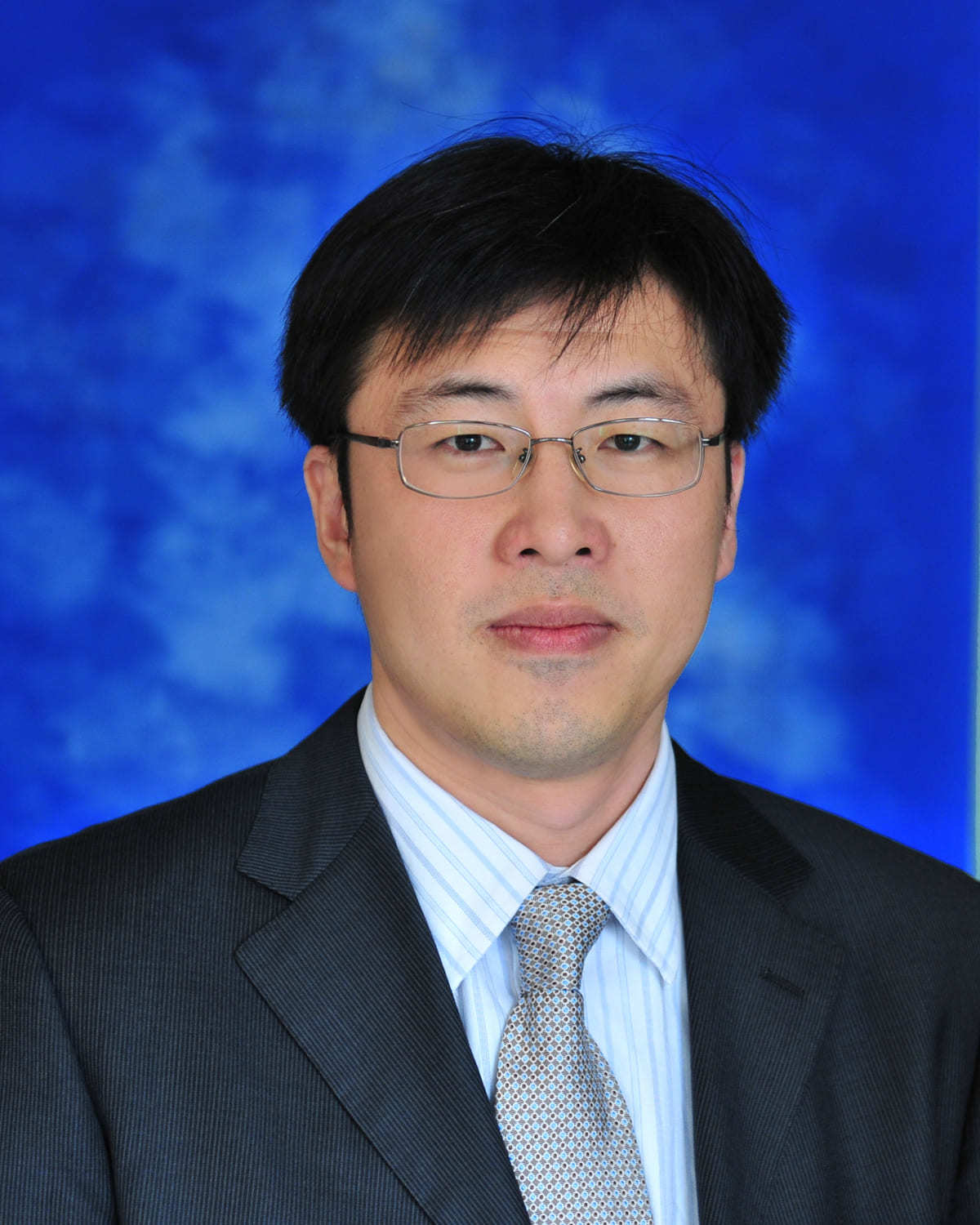}}]
	{Y. W. Peter Hong} (Senior Member, IEEE)  received
the B.S. degree in electrical engineering from
National Taiwan University, Taipei, Taiwan, in
1999, and the Ph.D. degree in electrical engineering
from Cornell University, Ithaca, NY, USA, in
2005. He joined the Institute of Communications
Engineering and the Department of Electrical
Engineering, National Tsing Hua University
(NTHU), Hsinchu, Taiwan, in 2005, where he is
a Full Professor. His research interests include AI/ML in wireless communications, UAV communications, and distributed signal
processing for IoT and sensor networks.

Dr. Hong received the IEEE ComSoc
Asia-Pacific Outstanding Young Researcher Award in 2010, the Y. Z. Hsu
Scientific Paper Award in 2011, the National Science Council Wu Ta-You
Memorial Award in 2011, the Chinese Institute of Electrical Engineering
Outstanding Young Electrical Engineer Award in 2012, the Best Paper
Award from the Asia-Pacific Signal and Information Processing Association
Annual Summit and Conference in 2013, and the Ministry of Science and
Technology Outstanding Research Award in 2018 and 2022, respectively. He currently serves as Senior Area Editor for IEEE TRANSACTIONS ON SIGNAL PROCESSING. He also previously served as an Associate
Editor for IEEE TRANSACTIONS ON SIGNAL PROCESSING and
IEEE TRANSACTIONS ON INFORMATION FORENSICS AND SECURITY, and
an Editor for IEEE TRANSACTIONS ON COMMUNICATIONS.
He was Chair of the IEEE ComSoc Taipei Chapter in 2017-2018, and
Co-Chair of the Technical Affairs Committee, the Information Services
Committee, and the Chapter Coordination Committee of the IEEE ComSoc
Asia-Pacific Board in 2014-2015, 2016-2019, and 2020-2021, respectively. He is currently Vice-Director of the IEEE ComSoc
Asia-Pacific Board and an IEEE ComSoc Distinguished Lecturer.
\end{IEEEbiography}

%\iffalse
\clearpage

\appendix
\section*{Appendix A}

\setcounter{section}{1}

In Table I, we provide a list of notations used in this paper.
%In Table \ref{tab:one}, we provide a list of notations used in this paper.

{
	\tiny
	\setlength{\tabcolsep}{10pt}
	\renewcommand\arraystretch{0.4}
	\begin{table}[ht]
	%	\begin{center}
			\caption{List of Notations\label{tab:one}}
			{%
				\scalebox{0.9}{
				\begin{tabular}{||l|l||}
					
				\hline\hline
               ${\cal A}$                      & A set of tokens (symbols)\\
			$c$                      		& The average number of skills in a (basic) text (see (A1))\\
			$c_{f}$                      		& The average number of skills in a domain-specific text (see (A1d))\\
			$c_{k,j}$                      		& The average number of class $k$ skills in a class $j$ text (see (A1m))\\
			${\tilde c}_{j}$                 & The vector of $c_{k,j}$, ${\tilde c}_{j} = ( c_{1,j}, c_{2,j}, \ldots, c_{K,j})$\\
                $\calD$                                 & The set of training texts\\
                $\calD_f$                            & The set of domain-specific texts for training\\
                $\calD_j$                           & The set of class $j$ training texts\\
			 $\tilde \calD_f$                            & The set of {\em learnable} domain-specific texts\\
$d_{k,j}$                        & The average number of class $j$ texts connected to a class $k$ skill node \\
			$d_{k}$                        &The mean degree of a class $k$ skill node \\
			$E$                                 & The edge set of the skill-text bipartite graph\\
%where an edges are between a skill $s$ and a text $t$ if $s$ is present in $\phi(t)$\\
                $G$                     & The skill-text bipartite graph\\
                $i$                                 & The number of SCNS iteration \\
                $J$                                 & The number of classes of texts \\
                $j$                                 & The index for the $j^{th}$ class of texts\\
                $K$                                 & The number of classes of skills \\
                $k$                                 & The index for the $k^{th}$ class of skills\\
                ${\cal L}$                        & A {\em semantic language}\\
                $M(s)$ & The number of edges connected to a skill node $s$ \\
                $M_{k,j}(s)$ & The number of class $(k,j)$-edges connected to a class $k$ skill node \\
                $N(t)$ & The number of edges connected to a text node $t$ \\
                $N_f(t)$ & The number of edges connected to a domain-specific text node $t$ \\
                $\tilde N_f(t)$ & The number of edges connected to a {\em learnable} domain-specific text node $t$ \\
                $N_{k,j}(t)$ & The number of class $(k,j)$-edges connected to a class $j$ text node \\
                $p_f$                            & The probability that the text end of a randomly selected edge in the learnable domain-specific bipartite graph is not learned after fine tuning\\
                $p_G$                        &The probability that a randomly selected skill node is in the giant component\\
                $p^{(i)}$                           & The probability that the text end of a randomly selected edge is not learned after the $i^{th}$ iteration\\
                $p_k^{(i)}$                    & The probability that the {\em text end} of a randomly selected class $k$ edge has not been learned after the $i^{th}$ SCNS iteration \\
			    $p_{k,j}^{(i)}$              & The probability that the {\em text end} of a randomly selected class $(k,j)$-edge has not been learned after the $i^{th}$ SCNS iteration\\
                $P_{\rm suc}(\rho)$              & The success probability of a Poisson learner when the number of skills in a text is Poisson distributed with mean $\rho$\\
                   $P_{\text{suc},k}(\rho)$            & The probability that a class $k$ skill is learned when the number of skills in a text is subject to a Poisson offered load $\rho$\\
                $\tilde P_{{\rm suc},k}^{(i)}$      & The probability that a randomly selected {\em class $k$ skill} can be learned after the $i^{th}$ SCNS iteration\\
%                $p_t$                            & The probability of testing errors of a randomly selected text from the domain-specific texts\\
                $q^{(i)}$                        & The probability that the skill end of a randomly selected edge is not learned after the $i^{th}$ SCNS iteration\\
			$q_k^{(i)}$                   & The probability that the {\em skill end} of a randomly selected class $k$ edge has not been learned after the $i^{th}$ SCNS iteration\\
                $R$                               & The ratio of the number of training texts to the number of skills\\
                $R_f$                            &The ratio of the number of training domain-specific texts to the number of domain-specific skills\\
                $r_{k,j}$                        & The probability that a class $k$ edge is a class $(k,j)$-edge\\
                $\calS$                            & The set of skills\\
                $\calS_k$                            & The set of  class $k$ skills\\
                $\calS_f$                            & The set of domain-specific skills\\
                $\tilde \calS_f$                            & The set of {\em learnable} domain-specific skills\\
                $\calT$                            & The set of texts composed of sequences of tokens\\
                $\calT_j$                            & The set of class $j$ texts\\
                $\calT_f$                            & The set of domain-specific texts\\
                $\tilde \calT_f$                            & The set of {\em learnable} domain-specific texts\\
                ${\cal Z}^+$                        & The set of nonnegative integers\\
                $\alpha_j$                      & The ratio of the number of class $j$ training texts to the total number of training texts\\
                $\beta_j$                      & The ratio of the number of class $k$ skills to the total number of skills\\
                $\epsilon$  & The probability of testing error of a randomly selected text \\
                $\epsilon_f$ & The probability of testing error of a randomly selected domain-specific text \\
                $\Theta_k$                       & The throughput of class $k$ skills from a text with a Poisson offered load $\rho$\\
                $\Lambda_f$               &The generating function of the degree distribution of a domain-specific skill node \\
                $\Lambda_{k,\ell}$      & The probability that a class $k$ skill node has $\ell$ edges\\
                $\Lambda_k(x)$               & The generating function of the degree distribution of a class $k$ skill node \\
			$\Lambda_{k}^\prime(x)$             & The derivative of $\Lambda_k(x)$\\
			$\Lambda_{k}^\prime(1)$             & The mean degree of a class $k$ skill node\\
        		$\lambda_k(x)$                 & The generating function of the excess degree distribution of a class $k$ skill node\\
                $\mu_s$                        &The probability that a skill node is connected to a small component via one of its edges\\
                $\mu_t$                        &The probability that a text node is connected to a small component via one of its edges\\
			$\rho$                             & The Poisson offered load $\rho=(\rho_1, \ldots, \rho_K)$\\
                $\rho_\ell$               &The probability of the number of basic skills required by a domain-specific skill is $\ell$\\
                $\phi$                                 & A function that maps a text $t$ in $T$ to a set of skills\\
                $\psi$                        & A deterministic function that maps a $K$-vector $n=(n_1, n_2, \ldots, n_K)$ to the $K$-vector $(\psi_1(n), \psi_2(n), \ldots, \psi_K(n))$\\
                 $\zeta$                               & The probability that a randomly selected skill is learned \\
                  $\zeta_f$                               & The probability that a randomly selected domain-specific skill is learned \\
				\hline
               \hline
			\end{tabular}}}
			\label{table:notations}
%		\end{center}
	\end{table}
}

\end{document}